\documentclass[a4paper,conference]{IEEEtran}

\pdfoutput=1

\usepackage{amsmath}
\usepackage{amssymb}
\usepackage{mathtools}
\usepackage{xcolor}
\usepackage{graphicx}
\usepackage{booktabs}
\usepackage{multirow}
\usepackage{hhline}
\usepackage{bm}
\usepackage{bbm}
\usepackage{url}
\usepackage{cite}
\usepackage{siunitx}
\usepackage{braket}
\usepackage[caption=false]{subfig}

\DeclareMathOperator*{\argmax}{argmax}

\ifCLASSINFOpdf
\else
\fi
\begin{document}
%

\title{The Forchheim Image Database for\\ Camera Identification in the Wild}


\author{\IEEEauthorblockN{Benjamin Hadwiger}
\IEEEauthorblockA{School of Engineering\\
Friedrich-Alexander-Universit\"at\\
Erlangen-N\"urnberg, Germany\\
Email: benjamin.hadwiger@fau.de}
\and
\IEEEauthorblockN{Chrstian Riess}
\IEEEauthorblockA{School of Engineering\\
Friedrich-Alexander-Universit\"at\\
Erlangen-N\"urnberg, Germany\\
Email: christian.riess@fau.de}}


%


\maketitle

\begin{abstract}
	
Image provenance can represent crucial knowledge in criminal investigation and journalistic fact checking. In the last two decades, numerous algorithms have been proposed for obtaining information on the source camera and distribution history of an image. For a fair ranking of these techniques, it is important to rigorously assess their performance on practically relevant test cases. To this end, a number of datasets have been proposed. 
However, we argue that there is a gap in existing databases: to our knowledge, there is currently no dataset that simultaneously satisfies two goals, namely a) to cleanly separate scene content and forensic traces, and b) to support realistic post-processing like social media recompression.

In this work, we propose the \emph{Forchheim Image Database} (FODB) to close this gap. It consists of more than 23,000 images of 143 scenes by 27 smartphone cameras, and it allows to cleanly separate image content from forensic artifacts. Each image is provided in 6 different qualities: the original camera-native version, and five copies from social networks.
%
We demonstrate the usefulness of FODB in an evaluation of methods for camera identification. We report three findings. First, the recently proposed general-purpose EfficientNet remarkably outperforms several dedicated forensic CNNs both on clean and compressed images. Second, classifiers obtain a performance boost even on unknown post-processing after augmentation by artificial degradations. Third, FODB's clean separation of scene content and forensic traces imposes important, rigorous boundary conditions for algorithm benchmarking.

\end{abstract}


%
\IEEEpeerreviewmaketitle

\section{Introduction}

With the emergence of affordable smartphones, it became straightforward to record images and videos and to share them via social networks.
However, this opportunity can also be abused for unlawful purposes. For instance, multimedia samples can depict illicit content like CSEM/CSAM, they may violate copyright, or may be intentionally aimed at deceiving the viewer. In such cases, authorship and authenticity of multimedia items can be a central question for criminal prosecution.

This motivated researchers to develop numerous image forensics algorithms over the last two decades. Initial methods mostly model imaging artifacts~\cite{lukas2006digital, farid2009survey, kirchner2015forensic}. More recently, deep learning-based approaches~\cite{yang2020survey, huh2018fighting, zhou2018learning, cozzolino2019noiseprint, bondi2017first, rafi2020remnet, caldelli2017image, phan2018identifying, moreira2018image} achieve state-of-the-art results.
These techniques enable a forensic analyst to detect and localize manipulations~\cite{farid2009survey, huh2018fighting, zhou2018learning, cozzolino2019noiseprint}, and to identify the source device~\cite{kirchner2015forensic, yang2020survey, lukas2006digital, bondi2017first, rafi2020remnet} or distribution history of images or videos~\cite{caldelli2017image, phan2018identifying, moreira2018image}. In this work, we limit our focus on the latter two tasks on images.

\begin{figure}[!t]
	\centering
	\vspace{3pt}
	\includegraphics[width=0.325\linewidth]{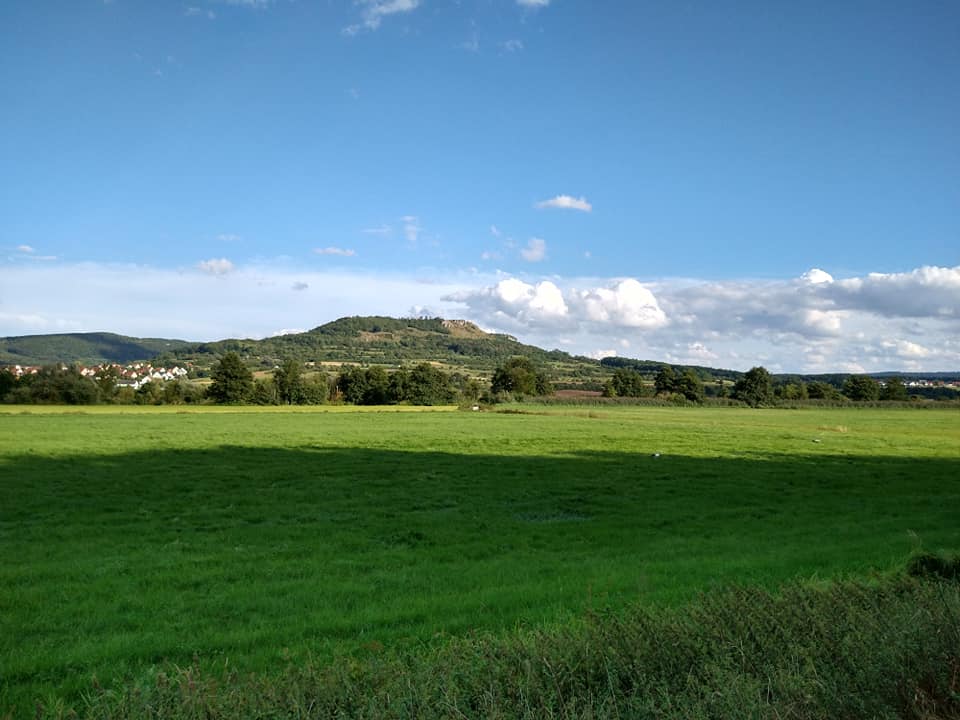}
	\includegraphics[width=0.325\linewidth]{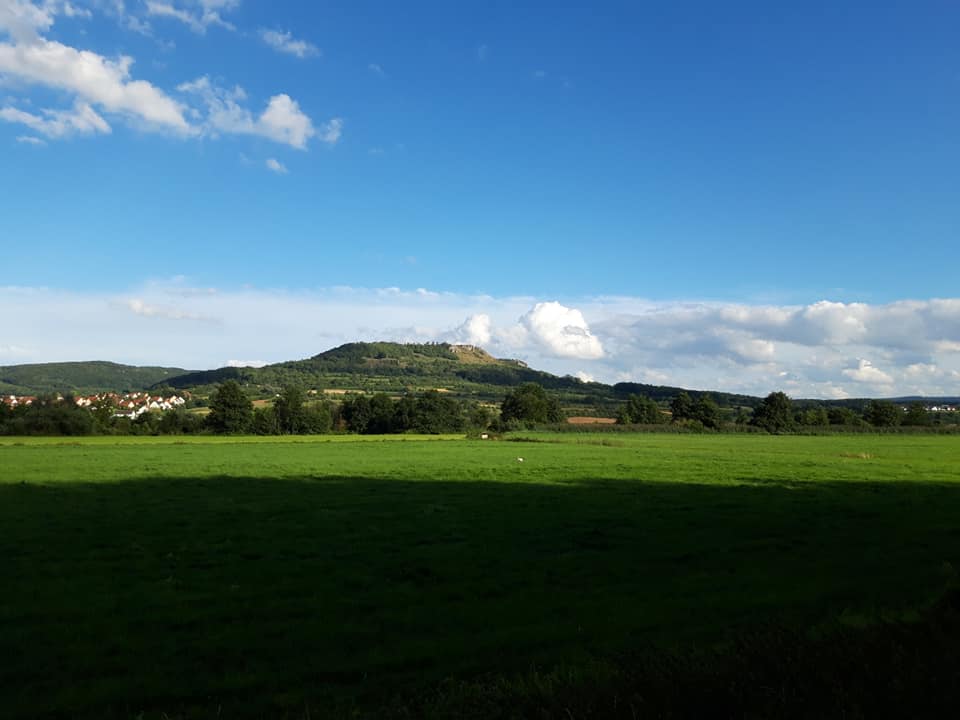}
	\includegraphics[width=0.325\linewidth]{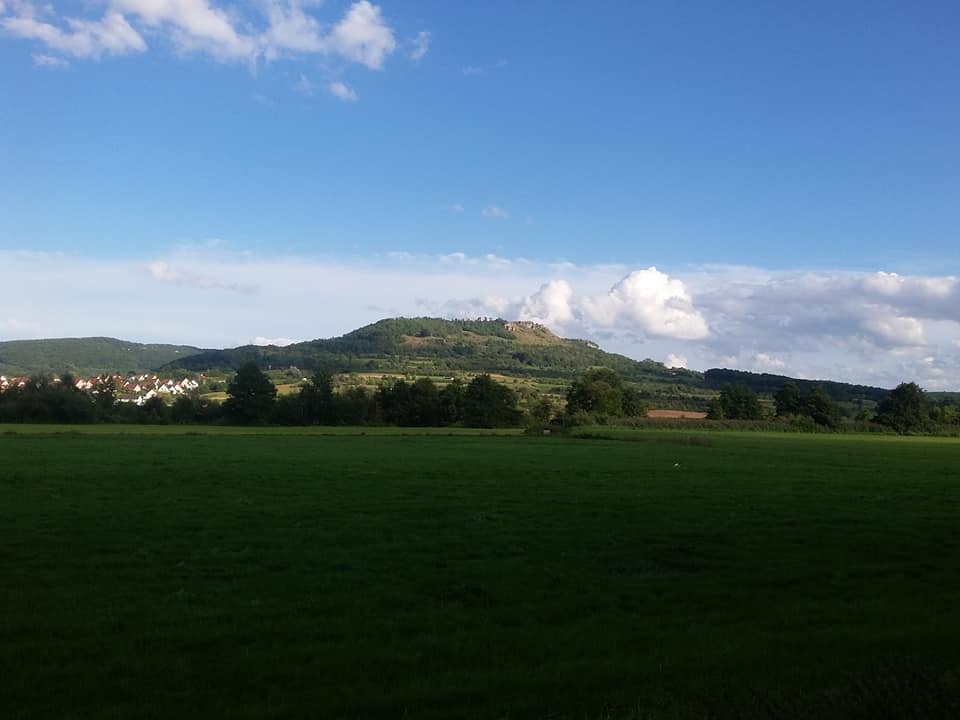}\\
	\vspace{3pt}
	\includegraphics[width=0.325\linewidth]{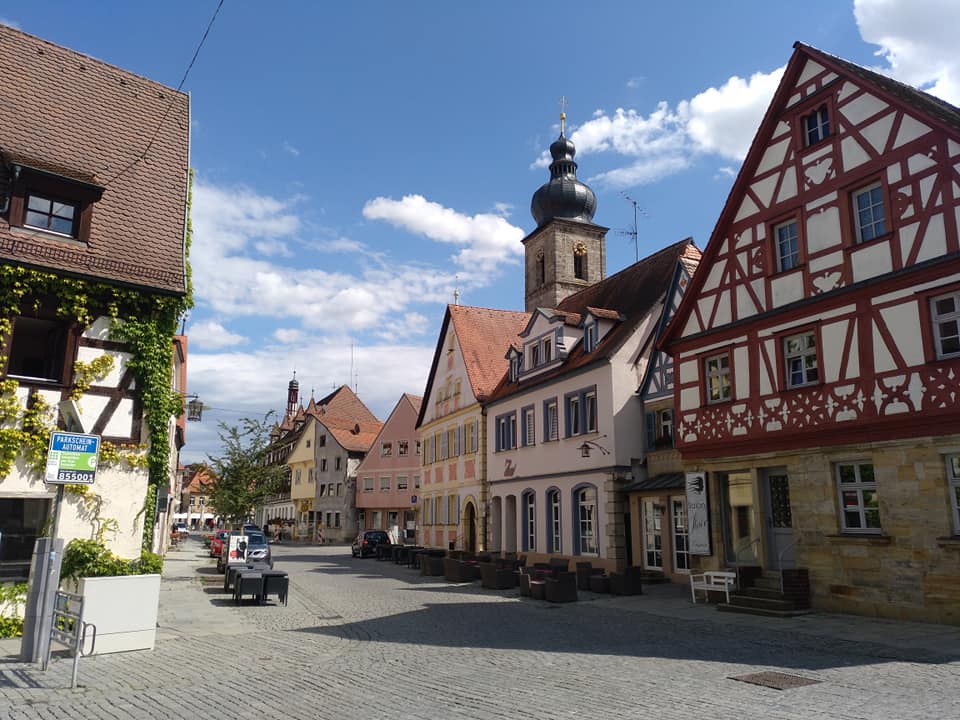}
	\includegraphics[width=0.325\linewidth]{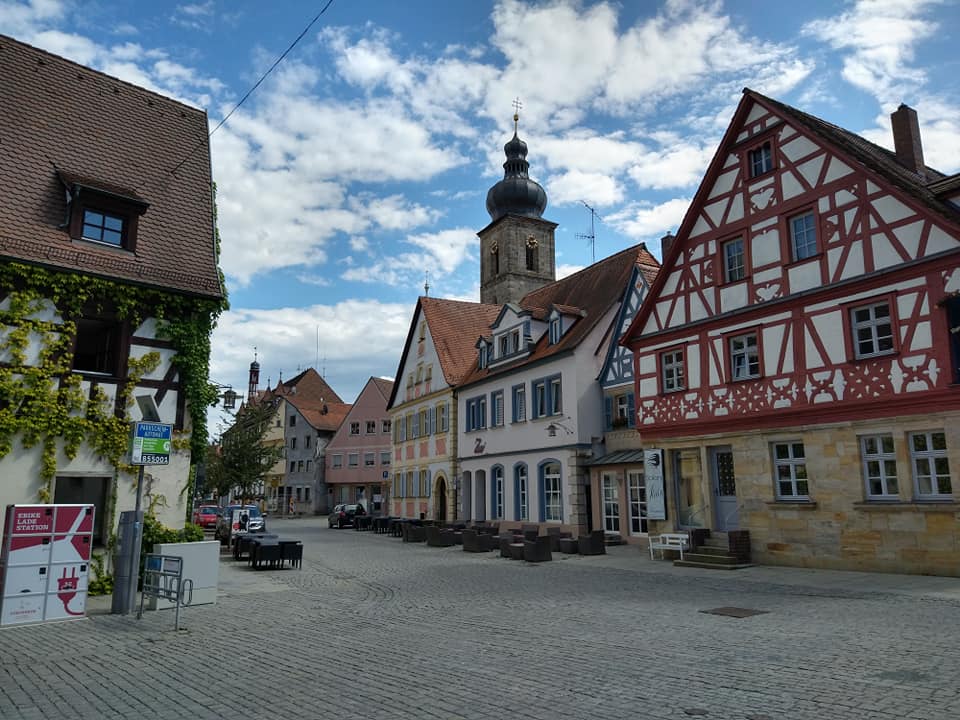}
	\includegraphics[width=0.325\linewidth]{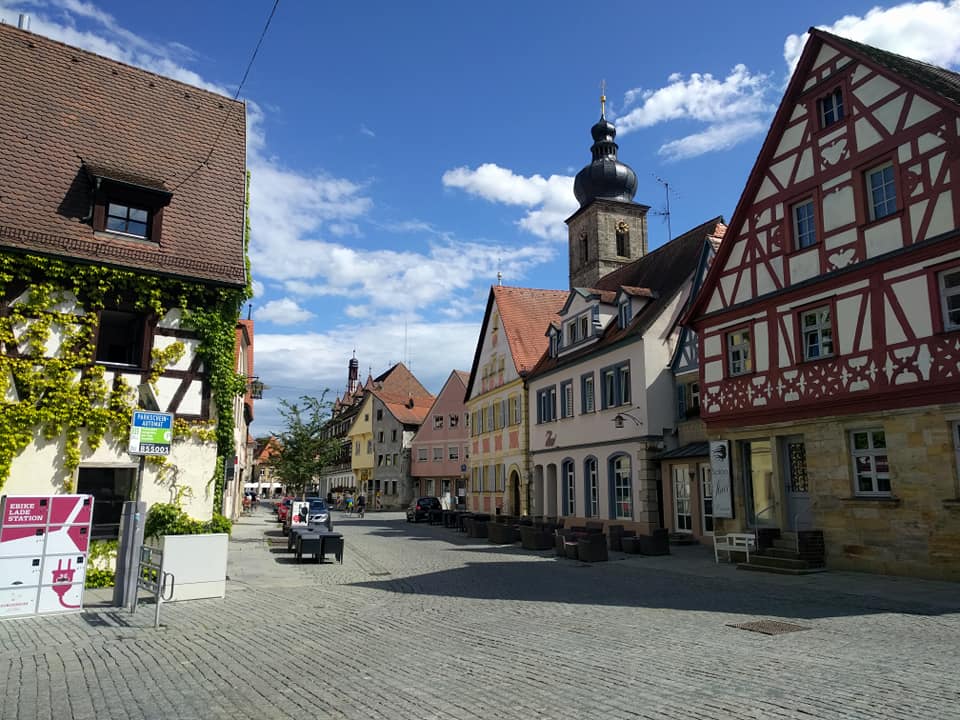}\\
	\vspace{3pt}
	\includegraphics[width=0.325\linewidth]{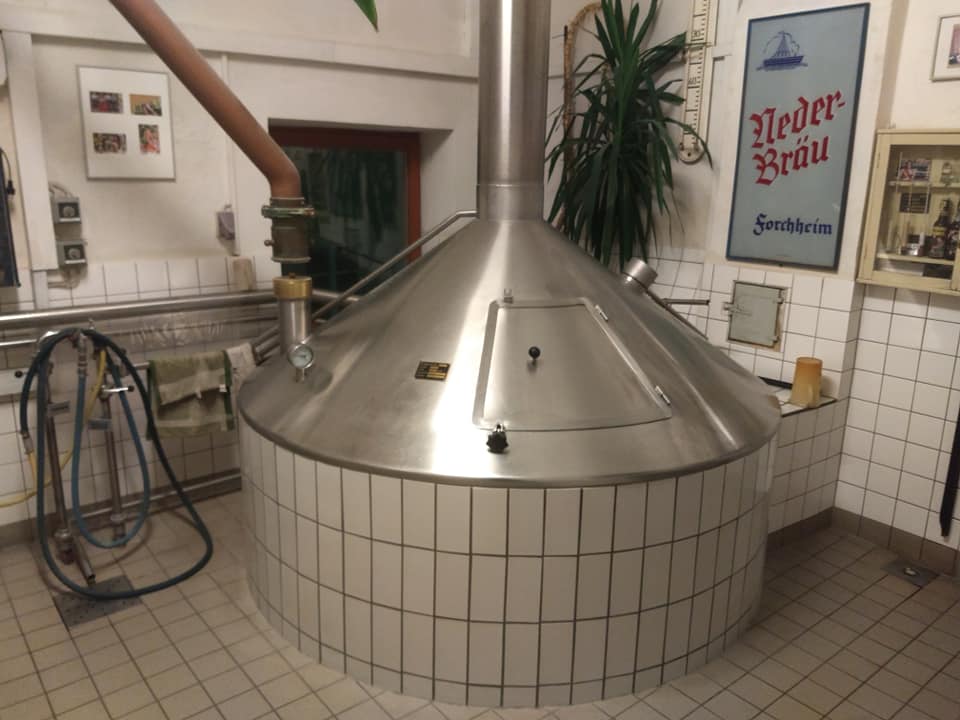}
	\includegraphics[width=0.325\linewidth]{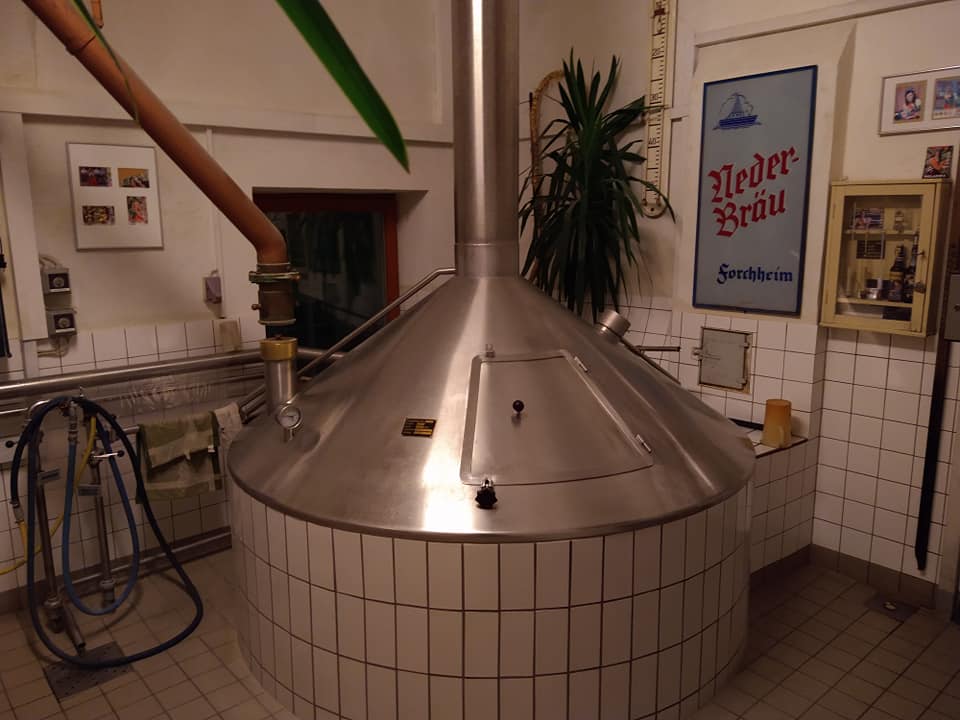}
	\includegraphics[width=0.325\linewidth]{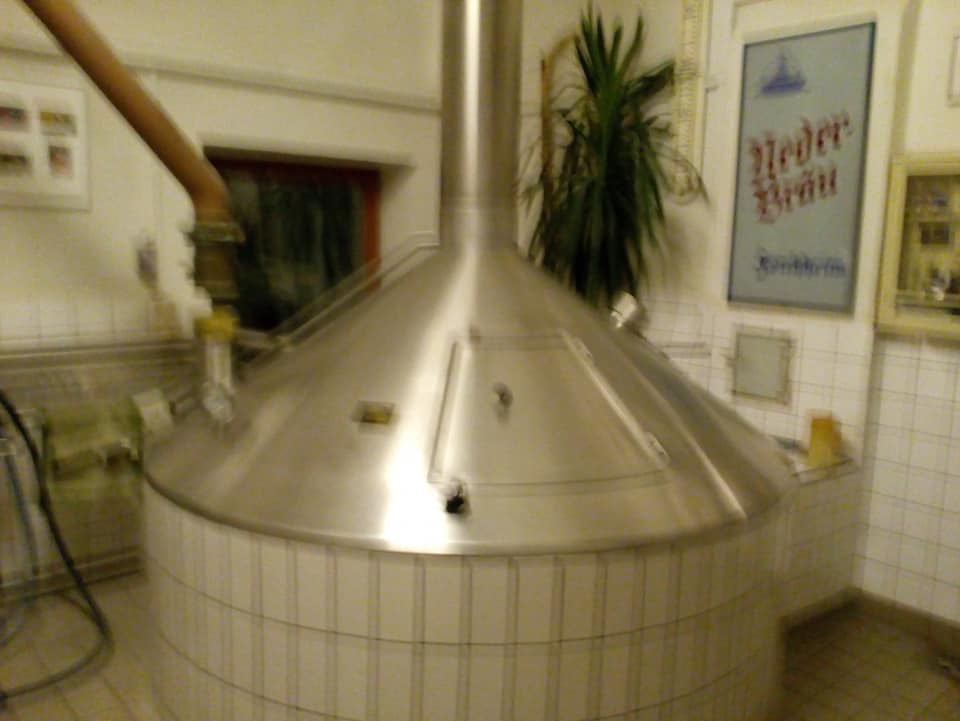}\\
	\vspace{3pt}
	\includegraphics[width=0.325\linewidth]{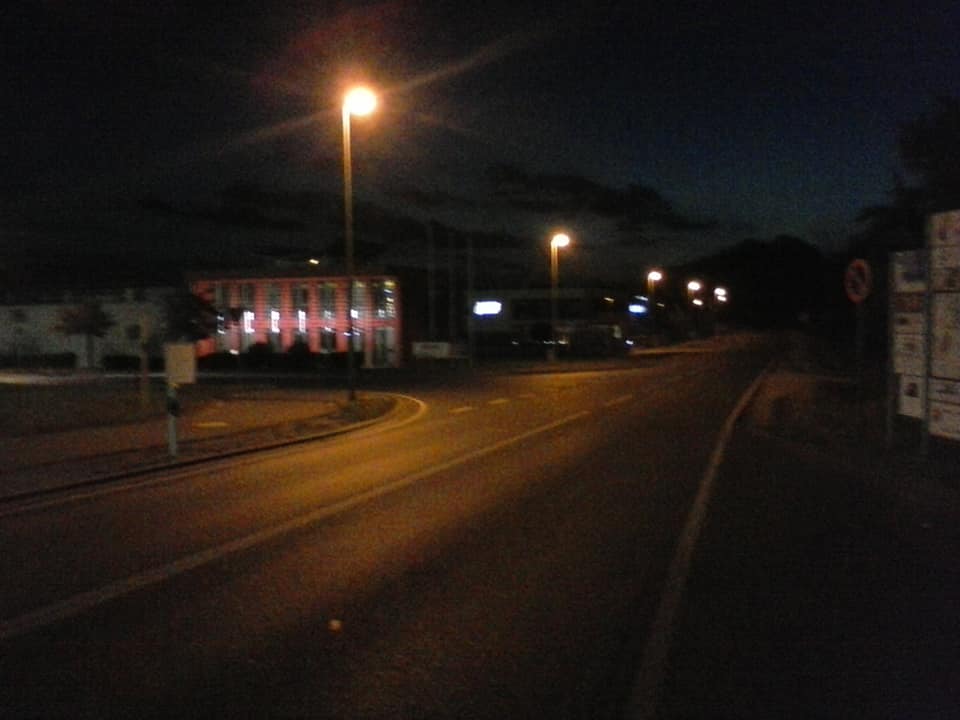}
    \includegraphics[width=0.325\linewidth]{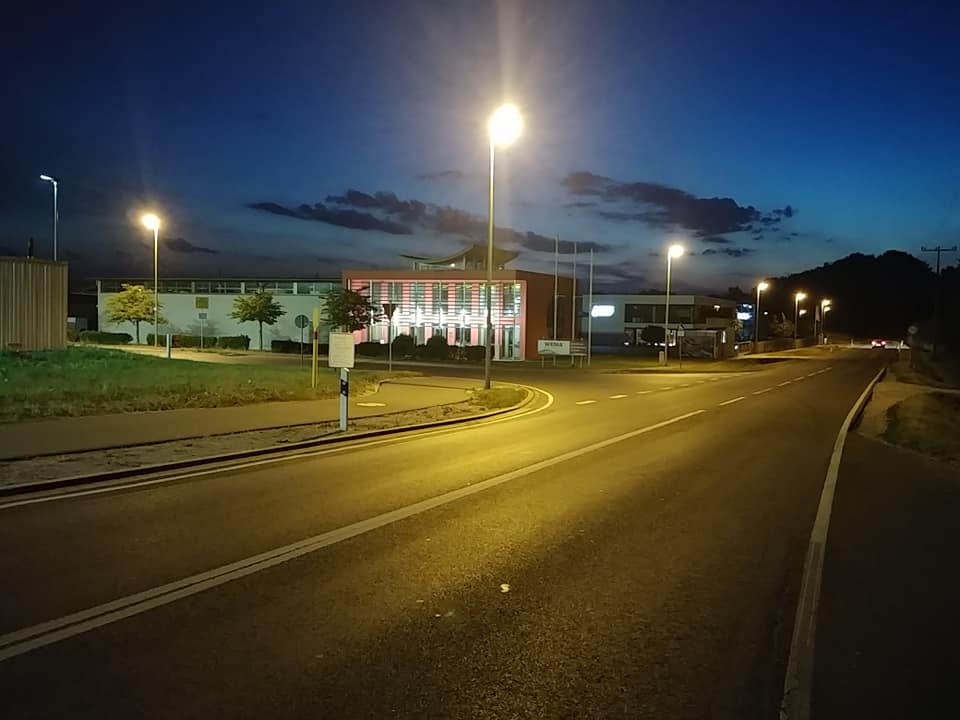}
	\includegraphics[width=0.325\linewidth]{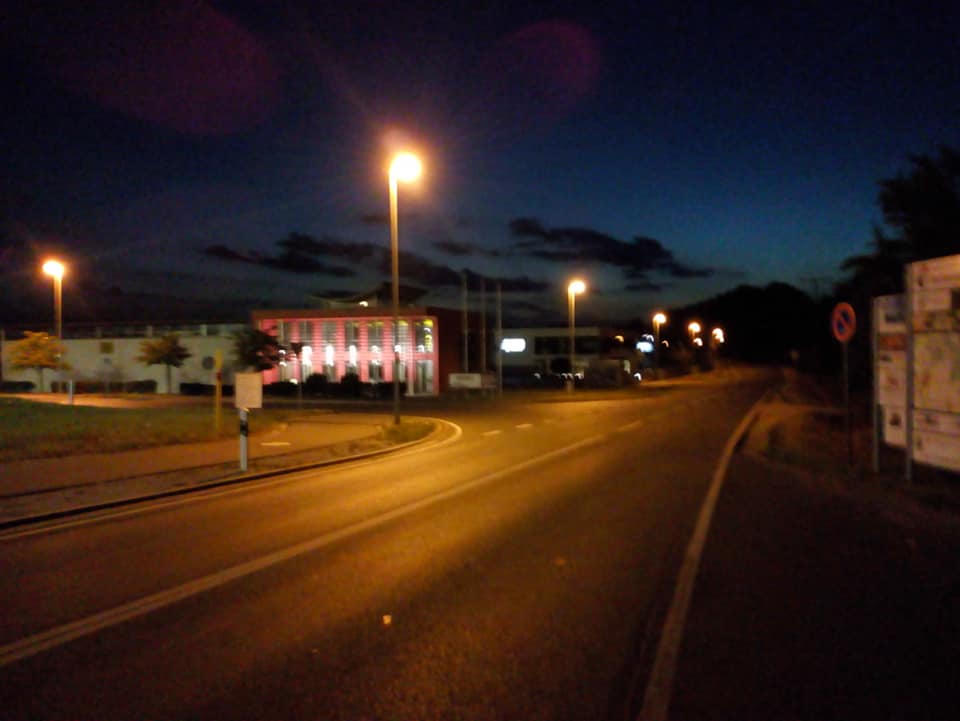}\\

	\caption{Example images from the Forchheim Image Database}
	\label{fig:example_images}
\end{figure}

The assessment of the real-world applicability of algorithms requires consistent evaluation protocols with standard benchmark datasets.
In 2010, Gloe and B\"ohme proposed the Dresden Image Database (DIDB)~\cite{gloe2010dresden}, the first large-scale benchmark for camera identification algorithms.
It consists of nearly 17,000 images of 73 devices depicting 83 scenes.
All devices record the same scenes. This is particularly important for aligning training/test splits with the scene content.
Doing so prevents the danger of opening a side channel through scene content, which may lead to overly optimistic results~\cite{kirchner2015forensic, bondi2017first}. 

The DIDB became in the past 10 years one of the most important benchmark datasets in the research community. However, it only consists of DSLR and compact cameras, whereas today most images are recorded with smartphones.
Also postprocessed versions of the images from social network sharing are not part of this dataset.

More recently, Shullani~\textit{et al.} proposed VISION~\cite{shullani2017vision}, an image and video database for benchmarking forensic algorithms.
It contains over 34,000 images in total, from 35 smartphones and tablet cameras. A subset of the images has been shared through Facebook and Whatsapp. This enables to investigate the impact of realistic post-processing on forensic traces. 

A limitation of VISION is that the images show arbitrary scenes. Thus, a training and evaluation split by scenes is not possible. Moreover, the scene content of images from the same camera are in some cases highly correlated.
This may be no issue for methods that strictly operate on noise residuals (e.g., PRNU-based fingerprinting~\cite{lukas2006digital}).
However, mixed scene content can open a side-channel for end-to-end Convolutional Neural Networks (CNNs), which potentially leads to overly optimistic evaluation results.

In this paper, we propose the Forchheim Image Database (FODB) as a new benchmark that combines the advantages of DIDB and VISION. It consists of 143 scenes, each captured with 27 smartphone cameras. Each image has been shared through the 5 social media apps by Facebook, Instagram, Telegram, Twitter, and Whatsapp. This yields a total of over 23,000 JPEG images. Examples from the database are shown in Fig.~\ref{fig:example_images}.
FODB allows training/test splits without scene overlap, and simultaneously supports robustness evaluations under real-world post-processing. Hence, it allows rigorous camera association benchmarking on real-world post-processing. 
To demonstrate the use of the dataset, we perform a benchmark of CNN-based camera identification, which brings insights into relative CNN performances, generalization to unseen post-processing, and performance impacts of scene splitting. In summary, our main contributions are:
\begin{itemize}
	\item We propose FODB, a new large-scale database for evaluating image forensics algorithms in the wild, which is available at \texttt{\url{https://faui1-files.cs.fau.de/public/mmsec/datasets/fodb/}}.
	\item We employ EfficientNet~\cite{tan2019efficientnet} for camera identification on FODB and 
	show that 
	it signficantly outperforms targeted forensic CNNs across almost all qualities.
	\item We show that degradation during training sigificantly boosts robustness even for unseen post-processing.
	\item We demonstrate the importance of scene splitting for learning-based camera identification
\end{itemize}

The remainder of the paper is organized as follows: We review image provenance benchmarks in Sec.~\ref{sec:related_work}. The proposed database FODB is described in Sec.~\ref{sec:database}. In Sec.~\ref{sec:method}, we describe our evaluation protocol for camera identification. The results of this evaluation are presented in Sec.~\ref{sec:results}. Section~\ref{sec:conclusion} concludes the paper.

\section{Related Work}
\label{sec:related_work}



In a number of existing datasets, different cameras replicate the same set of scenes. This allows to split the images into training and evaluation subsets such that scenes are disjoint. 

The first large-scale forensic benchmark to support such a splitting policy is the Dresden Image Database~\cite{gloe2010dresden}, as stated in the previous section.
Cheng~\textit{et al.} propose the NUS dataset~\cite{cheng2014illuminant}, with 1,736 images of over 200 scenes, each recorded with 8 DSLR cameras. In another work~\cite{cheng2015beyond}, Cheng~\textit{et al.} recorded additional 944 indoor images. Also in this dataset, each scene is captured with each camera.
Although the NUS dataset is presented as an illuminant estimation benchmark, it can directly be used for camera identification, and the acquisition protocols allow scene splitting similar to DIDB. Abdelhamed~\textit{et al.} propose the Smartphone Image Denoising Dataset (SIDD)~\cite{abdelhamed2018high} of about 30,000 images. It consists of 10 indoor scenes under different settings captured with 5 smartphone cameras.
The dataset targets image denoising, but can also be used for benchmarking
camera identification algorithms with proper scene splitting. 

Nowadays, images are often distributed via social networks and by that undergo compression to save memory and bandwidth. Therefore, it is important to assess the performance of forensic algorithms in the presence of such post-processing.
Unfortunately, social network sharing has not been relevant during conception
of these three datasets. Hence, none of these three datasets comes with images
that have already been passed through social networks. While a user of the
dataset could in principle pass the images through social networks by herself
(given permission by the creators of the datasets), it would still be a
remarkably tedious procedure. For example, we estimate that it would require
at least a month of work to upload and download the 17,000 images of DIDB
through various social networks due to limitations on automated image
uploading and downloading on most of their smartphone apps.

In 2018, the IEEE Signal Processing Society hosted a challenge for camera model identification~\cite{ieee2018camera}, which amongst other aspects addressed algorithm performance under general post-processing. The training dataset consists of 2,750 images of arbitrary scenes from 10 cameras.
The test dataset contains original images, as well as images that are recompressed with random JPEG quality, rescaling, or gamma correction.
In the VISION database by Shullani~\textit{et al.}, around 7,500 images of 35
smartphone cameras have been shared via Facebook in two qualities, and via
Whatsapp~\cite{shullani2017vision}. It consists of about 30,000 images in 4
qualities levels that enable evaluations of the impact of post-processing.
%
Guidice~\textit{et al.} propose a method for detecting the social network and
software used to share an image~\cite{giudice2017classification}.
To this end, they recorded images with 8 cameras of various types including 4 smartphones. Subsequently, they shared them via 10 social networks and two operating systems (OS) to obtain 2,720 images.
Caldelli~\textit{et al.} also investigate social network provenance~\cite{caldelli2017image}. They used 1,000 TIFF images from UCID~\cite{schaefer2003ucid}, an earlier image retrieval database. These images are compressed with different JPEG qualities and shared on 3 social networks, which results in 30,000 images. However, all images in UCID stem from a single camera, which does not allow for camera identification.
Phan~\textit{et al.} investigate traces of instant messenging apps and the host OS. They used 350 images out of 35 devices from the VISION dataset and shared them either once or twice with three messengers and two OSs~\cite{phan2018identifying}.
This leads to a total of 350 original images, 2,100 single-shared images and 6,300 double-shared images.
In a subsequent work,  Phan~\textit{et al.} consider up to three-fold sharing on social media platforms~\cite{phan2019tracking}.
They build two datasets. The first dataset is based on the raw image database RAISE~\cite{dang2015raise}.
The images are compressed in JPEG format and shared up to three times on three
social networks, which yields a total of 35,100 images.  The second dataset is
based one based on VISION. Here, 510 images are shared up to three times, to
obtain about additional 20,000 images.

The above stated datasets~\cite{ieee2018camera, shullani2017vision,giudice2017classification,caldelli2017image,phan2018identifying,phan2019tracking} allow benchmarking social network provenance algorithms. With the exception of the dataset by Caldelli~\emph{et al.} which consists of only one source camera~\cite{caldelli2017image},  they are also suitable for evaluating camera identification algorithms and their robustness for simulated~\cite{ieee2018camera} and real-world~\cite{shullani2017vision,giudice2017classification,phan2018identifying,phan2019tracking} post-processing. Two further large-scale camera identification benchmarks are SOCRatES~\cite{galdi2019socrates} and the Daxing Smartphone Identification Dataset (DSID)~\cite{tian2019daxing}. SOCRatES contains 9,700 images by 103 smartphones of 60 models, and thus is currently the database with largest number of devices. 
DSID consists of 43,400 images from 90 devices of 22 models, which currently is to our knowledge the database with the largest number of images and devices per model.

Unfortunately, none of these benchmark datasets supports scene splitting, such that it is currently not possible to investigate social media-related artifacts on split scenes.
However, we argue in line with previous works~\cite{kirchner2015forensic, bondi2017first}
that scene splitting is important during evaluation. It removes \emph{by design} the threat of leaking side-channel
information from the scene content into the evaluation. Such leakage may lead to an overestimation of the 
performance, as we will show in Sec.~\ref{sec:results}. The proposed Forchheim
Image Database FODB closes this gap: it jointly allows a rigorous scene
splitting policy, and enables to investigate the effect of social media
post-processing on forensic algorithms.

\section{The Forchheim Image Database}
\label{sec:database}

This section describes in detail the cameras, the acquisition protocol, the post-processing and database structure of the proposed dataset.
Table~\ref{tab:devices} lists the main features of the smartphones.
We use a total of of 27 smartphone devices, consisting of 25 different models from 9 brands.
It includes two models with more than one device, Samsung Galaxy A6 (devices 15
and 16) and Huawei P9 lite (devices 23 and 25). The smartphones run on Android
or iOS and represent older and more recent models (column ``Date'') with a wide range of retail
prices (not listed).
During image acquisition, we only use the main (i.e., rear) camera. All smartphones are configured to store images in JPEG format in the highest available resolution and the highest available JPEG quality. Focus, white-balance and High Dynamic Range (HDR) imaging is set to automatic mode, where applicable. 

All 143 scenes are captured in or near the town of Forchheim, Germany; hence the name Forchheim Image Database. Each camera recorded one image per scene. 
10 images are missing or excluded due to technical or privacy issues, resulting in a total of $3,861-10=3,851$ images. To assert diverse image content, we mix indoor and outdoor, day and night, close-up and distant, and natural and man-made scenes. Examples are shown in Fig.~\ref{fig:example_images}.

\begin{table}[!t]
	\centering
	\caption{Main features of smartphones in FODB}
	\label{tab:devices}
	\begin{tabular}{clllll}
		\toprule
		\textbf{ID} & \textbf{Brand} & \textbf{Model} & \textbf{OS} & \textbf{Version} & \textbf{Date}\\\midrule
		\textbf{01}& Motorola & E3 &  Android & 6.0 & 09/2016\\
		\textbf{02}& LG & Optimus L50 & Android & 4.4.2 & 06/2010\\
		\textbf{03}& Wiko & Lenny 2 & Android & 5.1 & 09/2014\\
		\textbf{04}& LG & G3 & Android & 5.0 & 07/2014\\
		\textbf{05}& Apple & iPhone 6s & iOS & 13.6 & 09/2015\\
		\textbf{06}& LG & G6 & Android & 9 & 05/2017\\
		\textbf{07}& Motorola & Z2 Play & Android & 8.0.0 & 08/2017\\ 
		\textbf{08}& Motorola & G8 Plus & Android & 9 & 10/2019\\
		\textbf{09}& Samsung & Galaxy S4 mini & Android & 4.4.4 & 05/2013\\ 
		\textbf{10} & Samsung & Galaxy J1 & Android & 4.4.4 & 01/2015\\
		\textbf{11} & Samsung & Galaxy J3 & Android & 5.1.1 & 01/2016\\
		\textbf{12} & Samsung & Galaxy Star 5280 & Android & 4.1.2 & 05/2013\\
		\textbf{13} & Sony & Xperia E5 & Android & 6.0 & 11/2016\\
		\textbf{14} & Apple & iPhone 3 & iOS & 7.1.2 & 06/2008\\
		\textbf{15} & Samsung & Galaxy A6 & Android & 10 & 05/2018\\
		\textbf{16} & Samsung & Galaxy A6 & Android & 10 & 05/2018\\
		\textbf{17} & Apple & iPhone 7 & iOS & 12.3.1 & 09/2016\\
		\textbf{18} & Samsung & Galaxy S4 & Android & 6.0.1 & 04/2013\\
		\textbf{19} & Apple & iPhone 8 Plus & iOS & 13.2 & 09/2017\\
		\textbf{20} & Google & Pixel 3 & Android & 9 & 11/2018\\
		\textbf{21} & Google & Nexus 5 & Android & 8.1.0 & 10/2015\\
		\textbf{22} & BQ & Aquaris X & Android & 8.1.0 & 05/2017\\ 
		\textbf{23} & Huawei & P9 lite & Android & 6.0 & 05/2016\\ 
		\textbf{24} & Huawei & P8 lite & Android & 5.0 & 04/2015\\ 
		\textbf{25} & Huawei & P9 lite & Android & 7.0 & 05/2016\\ 
		\textbf{26} & Huawei & P20 lite & Android & 8.0.0 & 04/2018\\
		\textbf{27} & Google & Pixel XL & Android & 10 & 10/2016\\
		\bottomrule
	\end{tabular} 
\end{table}

We refer to camera-native images as \emph{original} (orig.). Additionally, we created five post-processed versions of each image.
For this, we installed the apps Facebook, Instagram, Telegram, Twitter and Whatsapp on each device\footnote{Exceptions: Devices 2, 12, 14, 24, 25 did not support some apps, hence we transferred the images to other devices of the same OS (2, 12 $\rightarrow$ Device 8; 14 $\rightarrow$ Device 5; 24, 25 $\rightarrow$ Device 20) and shared all images from there.} and manually shared all images.
In the Facebook app, we uploaded the images of each device to a dedicated photo album in default quality\footnote{Corresponding to ``FBL'' (Facebook low quality) in the VISION database.}.
Then, we used the functionality to download entire albums in the browser version of Facebook.
During upload on Instagram, a user must select a square crop from an image, and optionally a filter.
We uploaded all images with default settings for cropping, resulting in a center crop, and disabled any filters.
For download we used the open source tool ``Instaloader'' (Version 4.5.2)\footnote{\url{https://instaloader.github.io/}}. In the Twitter app, all images were uploaded without filter, and downloaded via the Firefox browser plugin ``Twitter Media Downloader'' (Version 0.1.4.16)\footnote{\url{https://addons.mozilla.org/de/firefox/addon/tw-media-downloader/}}. For Telegram and Whatsapp, the images of each device were sent to device 6 (LG G6), except for the images of device 6 itself, which were sent to device 8 (Motorola G8 Plus). In this way, the database contains a total of $6\cdot (27\cdot 143 - 10)= \num{23106}$ JPEG images.

Social network and messenger sharing was executed one device after another, to avoid confounding images of different devices.
During sharing, the social networks and messengers non-trivially modify the image filenames, and metadata is largely removed. For re-identifying the shown scene, we correlated the original and post-processed images for each device individually. The originals were first downscaled to match the size of the post-processed versions, and, in case of Instagram, center cropped prior to downscaling. Only very few cases were ambiguouos, which were manually labeled.

The database is hierarchically organized: at root level, images from each device are in one directory  
\texttt{D$\langle$ID$\rangle$\_$\langle$Brand$\rangle$\_$\langle$Model$\rangle$\_$\langle$i$\rangle$}, where \texttt{ID}, \texttt{Brand} and \texttt{Model} are substituted according to Tab.~\ref{tab:devices}, and \texttt{i} $\in\{1,2\}$ enumerates the devices of a model.
Each directory contains six provenance subdirectories \texttt{orig}, \texttt{facebook}, \texttt{instagram}, \texttt{telegram}, \texttt{twitter} and \texttt{whatsapp}. These directories contain the images of device \texttt{ID}, provenance \texttt{prov} and scene ID \texttt{scene} with the pattern \texttt{D$\langle$ID$\rangle$\_img\_$\langle$prov$\rangle$\_$\langle$scene$\rangle$.jpg}, for example \texttt{D06\_img\_twitter\_0030.jpg}. 


\section{Camera Identification: Dataset Split, Methods, and Training Augmentation}
\label{sec:method}
We demonstrate an application of FODB by studying the behavior of CNNs for
camera identification.  This section briefly describes the used methods and their training.

\subsection{Dataset Splits}
To create training, validation and test data, we split the set of 143 scenes
$\mathcal{S}$ of FODB into three disjoint sets $\mathcal{S}^\mathrm{train}$,
$\mathcal{S}^\mathrm{val}$ and $\mathcal{S}^\mathrm{test}$, and we set
$\lvert\mathcal{S}^\mathrm{train}\vert=97$,
$\lvert\mathcal{S}^\mathrm{val}\vert=18$,
$\lvert\mathcal{S}^\mathrm{test}\vert=28$. For camera models with more than one
device, we choose the device with the smallest \texttt{ID}, which yields
$N_\mathcal{K} = 25$ cameras, and hence $25$ classes. Thus we obtain
$\lvert\mathcal{S}^\mathrm{train}\rvert \cdot N_\mathcal{K} = 2425$ training,
$\lvert\mathcal{S}^\mathrm{val}\rvert \cdot N_\mathcal{K} = 450$ validation and
$\lvert\mathcal{S}^\mathrm{test}\rvert \cdot N_\mathcal{K} = 700$ test images
per post-processing quality. 

\subsection{Compared Methods}
We reimplemented three CNN-based forensic methods for source camera
identification. First, the method by Bondi~\textit{et al.}, which we
subsequently refer to as ``BondiNet''~\cite{bondi2017first}. Second, MISLnet by
Bayar~\textit{et al.}~\cite{bayar2018constrained} in its improved version as
the feature extractor in the work by Mayer~\emph{et
al.}~\cite{mayer2019forensic} by the same group. Third, RemNet by
Rafi~\textit{et al.}~\cite{rafi2020remnet}, which has been presented at the
IEEE Signal Processing Cup 2018. We additionally report results on
EfficientNet-B5, a recently proposed general-purpose CNN from the field of
computer vision~\cite{tan2019efficientnet}. All models are trained with crossentropy loss.

The input patch size of each CNN except MISLNet is set to $64\times 64$ pixels.
The outputs of the CNNs are adapted to distinguish $N_\mathcal{K}=25$ classes.
Note that the classes are balanced, and random guessing accuracy is 
$N_\mathcal{K}^{-1}$, i.e., $4\%$, on all experiments on FODB.

Initial experiments with BondiNet using the parameters of the
paper~\cite{bondi2017first} led to poor validation performance on FODB. Hence,
we evaluate BondiNet for the following set of hyperparameters, which led to
significantly better validation results: Adam optimizer with $\alpha=10^{-3}$,
$\beta_1=0.9$ and $\beta_2=0.999$, no weight decay, additional batch
normalization after each convolution, and direct classification using CNN
outputs instead of using an SVM.
For MISLnet, we reimplemented the improved version of the same
group~\cite{mayer2019forensic}. The patch input size is $256\times 256$ pixels,
and hence somewhat larger than for the remaining networks. We address this in
the patch clustering described below.  For RemNet, we reimplemented the
implementation as described in the paper.

For EfficientNet-B5, we use weights pretrained on
ImageNet~\cite{deng2009imagenet}, and remove the final classification layer of
the pretrained network. Then, we add global average pooling and a dense layer
with $N_\mathcal{K}=25$ output units and softmax activation. The weights of the new
classification layer are set with Glorot uniform
initialization~\cite{glorot2010understanding}. During all experiments, we use
Adam optimization~\cite{kingma2015adam} with learning rate $\alpha=10^{-3}$ and
moments $\beta_1=0.9$ and $\beta_2=0.999$. Whenever the validation loss
stagnates for two consecutive epochs, the learning rate is halved, and we apply
early stopping. 

To accomodate for differences in the input resolution of the networks,
we adopt the \emph{patch cluster} strategy by Rafi~\textit{et
al.}~\cite{rafi2020remnet}. To this end, we consider an image area of
$256\times 256$ pixels as a patch cluster. A patch cluster is considered to be
non-informative if it mostly consists of homogeneous pixels, which is
determined by a heuristic quality criterion used by Bondi~\emph{et
al.}~\cite[Eqn.~(1)]{bondi2017preliminary} and Rafi~\emph{et
al.}~\cite[Eqn.~(7)]{rafi2020remnet},
\begin{equation}
Q(P) = \frac{1}{3} \sum\limits_{c \in \{R, G, B\}} \alpha\beta (\mathbf{\mu}_c - \mathbf{\mu}_c^2) + (1-\alpha) (1 - \exp(\gamma \sigma_c))\enspace,
\end{equation}
where $\mathbf{\mu}_c$ and $\sigma_c$ denote the patch cluster mean and standard
deviation in the red, green, and blue color channels $c$, and $\alpha = 0.7$,
$\beta = 4$, $\gamma = \ln(0.01)$.

\subsection{Matching the Network Input Resolutions}
For evaluation, it is important to provide to each algorithm the same amount of
information. Thus, for both training and testing, we subdivide the image into
non-overlapping $256\times 256$ patch cluster candidates, and sort them by the
quality criterion $Q(P)$. The top $100$ candidates are used as patch clusters for
the image.

For training, each selected patch cluster is used once per training epoch.
MISLnet obtains full patch clusters to match its $256\times 256$ pixel inputs.
The remaining networks obtain a randomly selected $64\times 64$ pixels
subwindow per cluster to match their input size. For validation, we use the
same approach but with fixed random seed to achieve a consistent comparison
throughout all epochs.

For testing, we also feed a full $256\times 256$ patch cluster to MISLnet. For
the remaining networks, we subdivide a patch cluster into $16$ non-overlapping
patches of $64\times 64$ pixels.

These results are used to report three metrics: camera identification
accuracies on individual $64\times 64$ patches (excluding MISLnet),
accuracies on patch clusters of $256\times 256$ pixels, and accuracies for
the whole image. For the per-cluster accuracies, we directly calculate the
accuracy for MISLnet's $256\times 256$ prediction. For the remaining networks,
the patch cluster prediction $\hat{k}$ is calculated via
soft majority voting over all $64\times 64$ patch predictions,
\begin{equation}
\label{eq:majority_voting}
	\hat{k} = 
	\argmax\limits_k\sum\limits_{i\in\mathcal{I}} y_i^k\enspace,
\end{equation}
where $y_i^k$ denotes the $k$-th component of the CNN output for the $i$-th patch 
in the cluster. The prediction for the whole image is analogously calculated via soft 
majority voting over all individual CNN outputs on that image.

\subsection{Training Augmentation}\label{sec:augmentation}
Throughout all training runs, we randomly apply horizontal flipping, vertical flipping,
and rotation by integer multiples of $90^\circ$, with equal probability for
each case.

For a subset of the experiments, we additionally apply artificial degradations during
training to increase the robustness of all CNNs against post-processing.
Prior to flipping or rotation, we rescale a training patch cluster with
probability $0.9$. The rescaling factor is randomly drawn from a
discrete distribution over the interval $[0.25, \ldots, 4]$. In order to make
upsampling and downsampling equally likely, we rewrite the interval as $[0.25
= 2^{-2}, \ldots, 2^j, \ldots, 4 = 2^2]$ and subdivide the exponent $j$ in
$50$ equally spaced samples. We draw from these exponents with uniform
probability.
After flipping or rotation,  we extract a patch from the (rescaled or
non-rescaled) cluster, and recompress it in JPEG format with probability $0.9$.
The JPEG quality factor is uniformly chosen from $[100, 90, \dots,10]$.

For the rather challenging experiments in Sec.~\ref{sec:realworld} and
Sec.~\ref{sec:splitting}, we try to maximize the performance of RemNet and
EfficientNet-B5 by pretraining on DIDB's 18 cameras. To this end, we use the
DIDB training/validation split by Bondi~\emph{et al.}~\cite{bondi2017first}.
Considering our four variants of RemNet and EfficientNet-B5 with and without
artificial degradations, we investigate possible gains in validation
performance when pre-training on DIDB. We apply artificial degradations on DIDB
only if the subsequent 
training on FODB also uses artificial
degradations, in order to have these degradations either throughout the whole
training process or not at all.  We then select for the experiments either the
variant with DIDB-pretraining or without depending
on the validation loss.  The results are listed in
Tab.~\ref{tab:modelselection}. Boldface shows validation loss and accuracy of
the selected model. The column indicating validation on FODB is used in
Sec.~\ref{sec:realworld}, the column indicating validation on VISION in
Sec.~\ref{sec:splitting}.

\begin{table}[!h]
	\centering
	\caption{Averaged overall validation performance for FODB and VISION, with
	and without pretraining on DIDB. Boldface shows the selected model variants
	based on validation loss.}
	\label{tab:modelselection}
	\begin{tabular}{ccccccc}
		\toprule
		\multicolumn{3}{c}{}&\multicolumn{4}{c}{\textbf{Validation Dataset}}\\\cmidrule{4-7}
		\multicolumn{3}{c}{\textbf{Training Parameters}} & \multicolumn{2}{c}{\textbf{FODB}}&\multicolumn{2}{c}{\textbf{VISION}}\\\midrule
		\textbf{Model}&\textbf{Degr.} & \textbf{Pretr.} & \textbf{Loss} & \textbf{Acc.} & \textbf{Loss} & \textbf{Acc.}\\\midrule
		\textbf{RemNet} & no & no &  \textbf{0.1870} & \textbf{92.72} & 0.1898 & 93.90 \\
		\textbf{RemNet} & no & yes & 0.1885 & 92.86 & \textbf{0.1731} & \textbf{94.49} \\\midrule
		\textbf{RemNet} & yes & no &  \textbf{2.4268} & \textbf{31.06} & 1.9586 & 42.67\\
		\textbf{RemNet} & yes & yes & 2.5735 & 26.07 & \textbf{1.9295} & \textbf{43.47}  \\\midrule
		\textbf{EN-B5} & no & no &  \textbf{0.1176} & \textbf{95.79} & 0.1465 & 95.91 \\
		\textbf{EN-B5} & no & yes & 0.1178 & 95.62 & \textbf{0.1265} & \textbf{96.22}\\\midrule
		\textbf{EN-B5} & yes & no & 1.6894 & 52.12 & 1.2410 & 63.68\\
		\textbf{EN-B5} & yes & yes & \textbf{1.6756} & \textbf{52.77} & \textbf{1.2179} & \textbf{64.35}\\\bottomrule
	\end{tabular}	
\end{table}

\begin{figure*}[!t]
	\centering
	\begin{subfloat}{\includegraphics[width=0.245\textwidth]{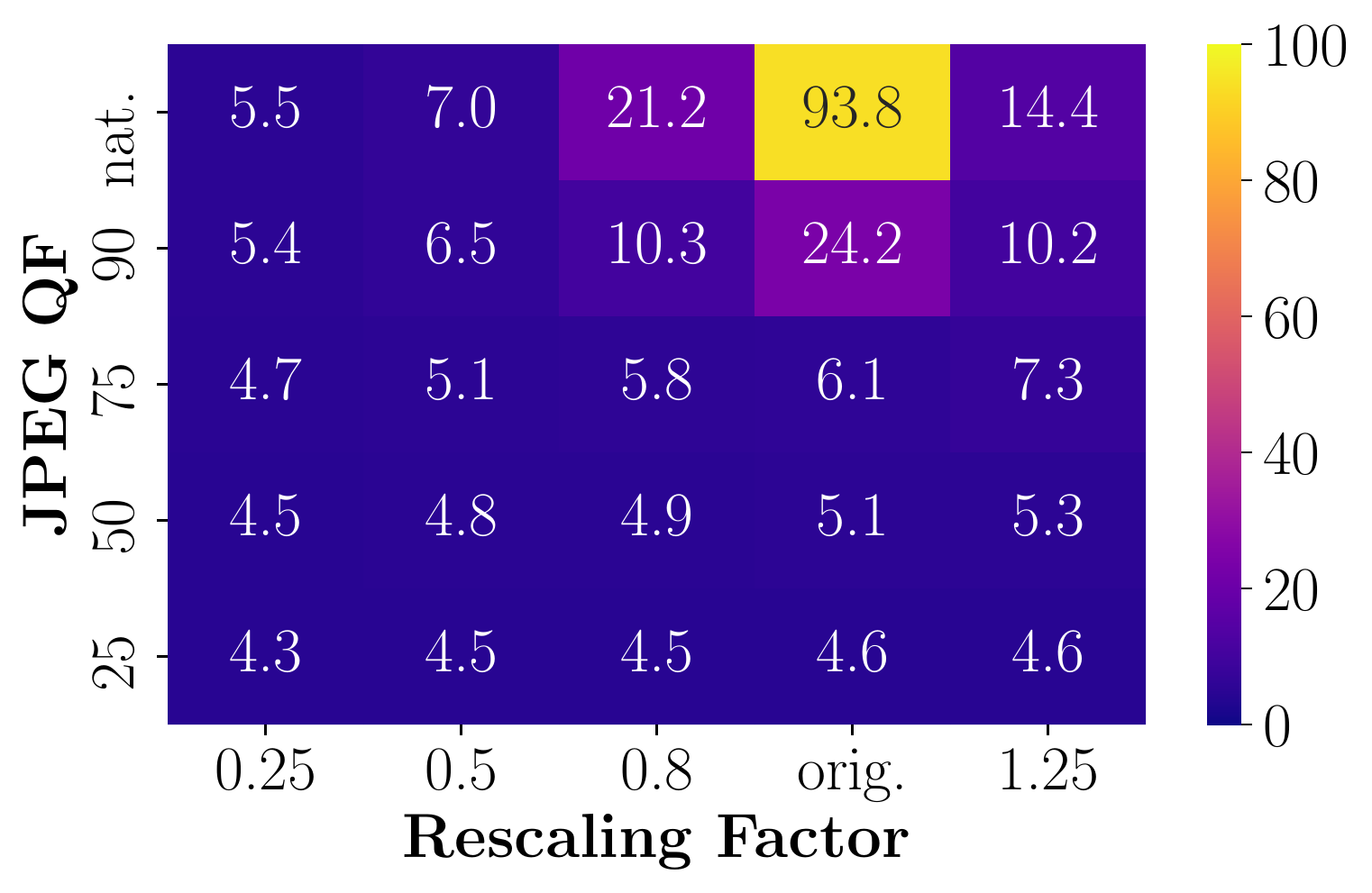}}
	\end{subfloat}
	\begin{subfloat}{\includegraphics[width=0.245\textwidth]{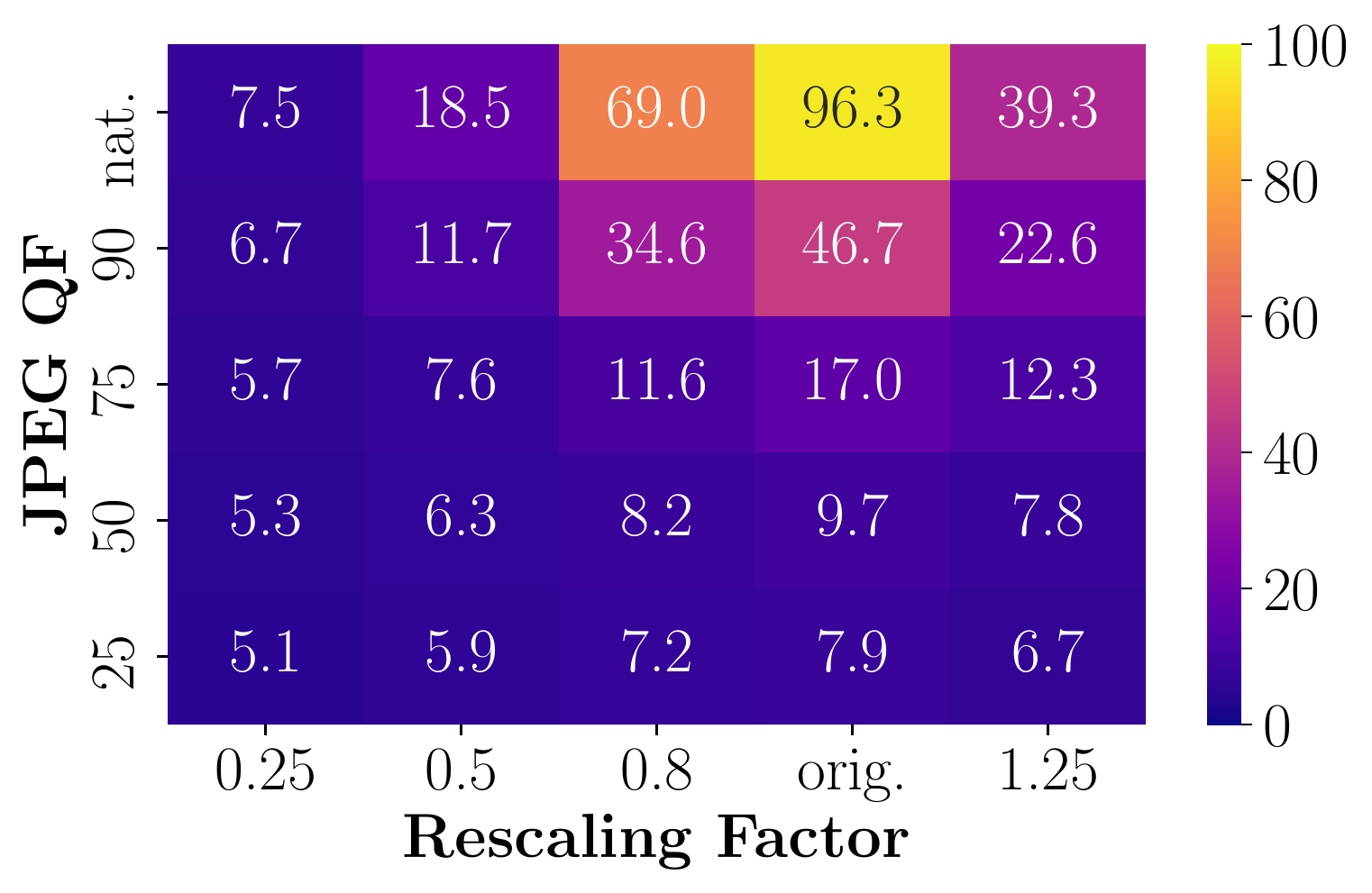}}
	\end{subfloat}
	\begin{subfloat}{\includegraphics[width=0.245\textwidth]{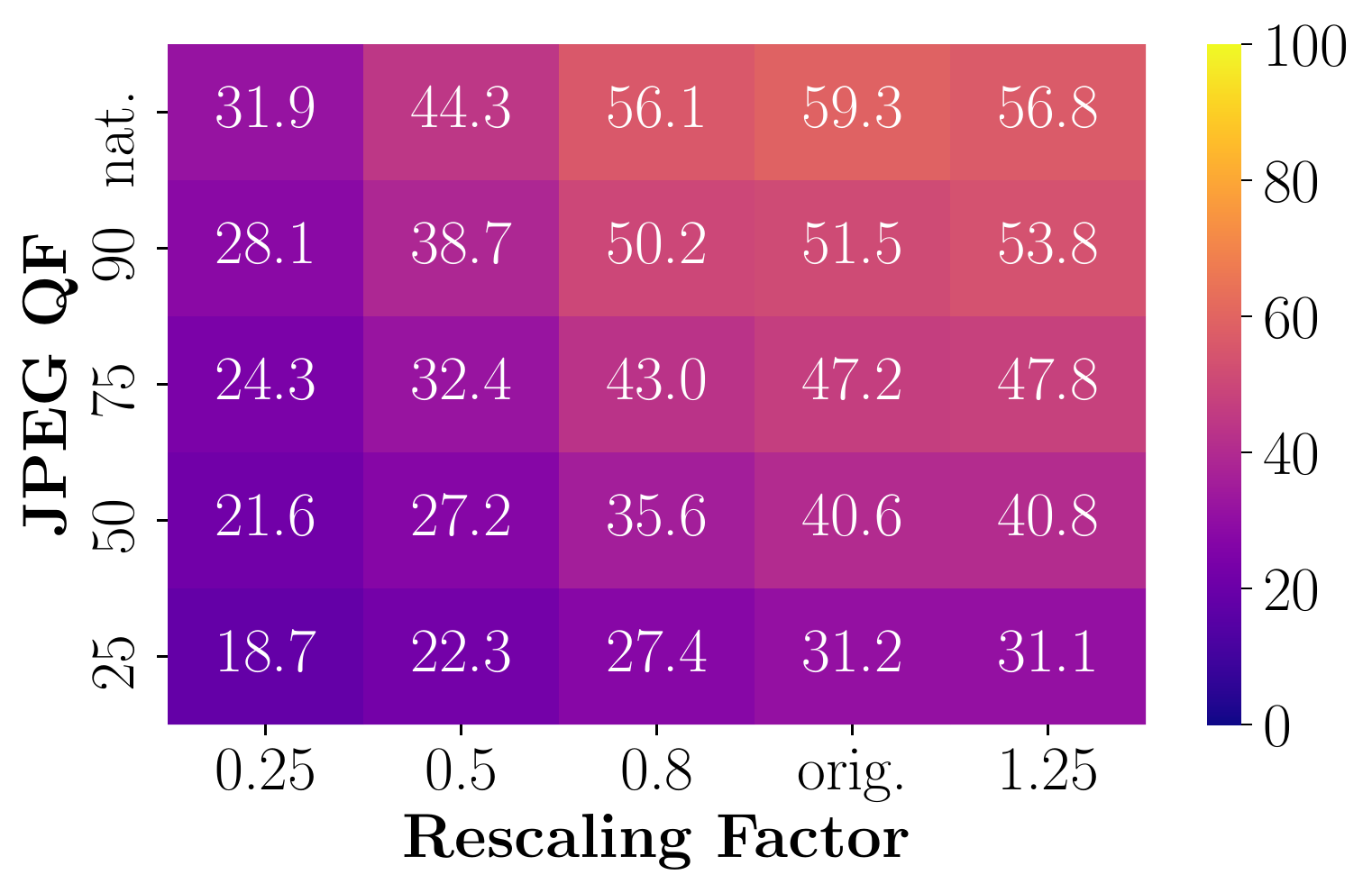}}
	\end{subfloat}
	\begin{subfloat}{\includegraphics[width=0.245\textwidth]{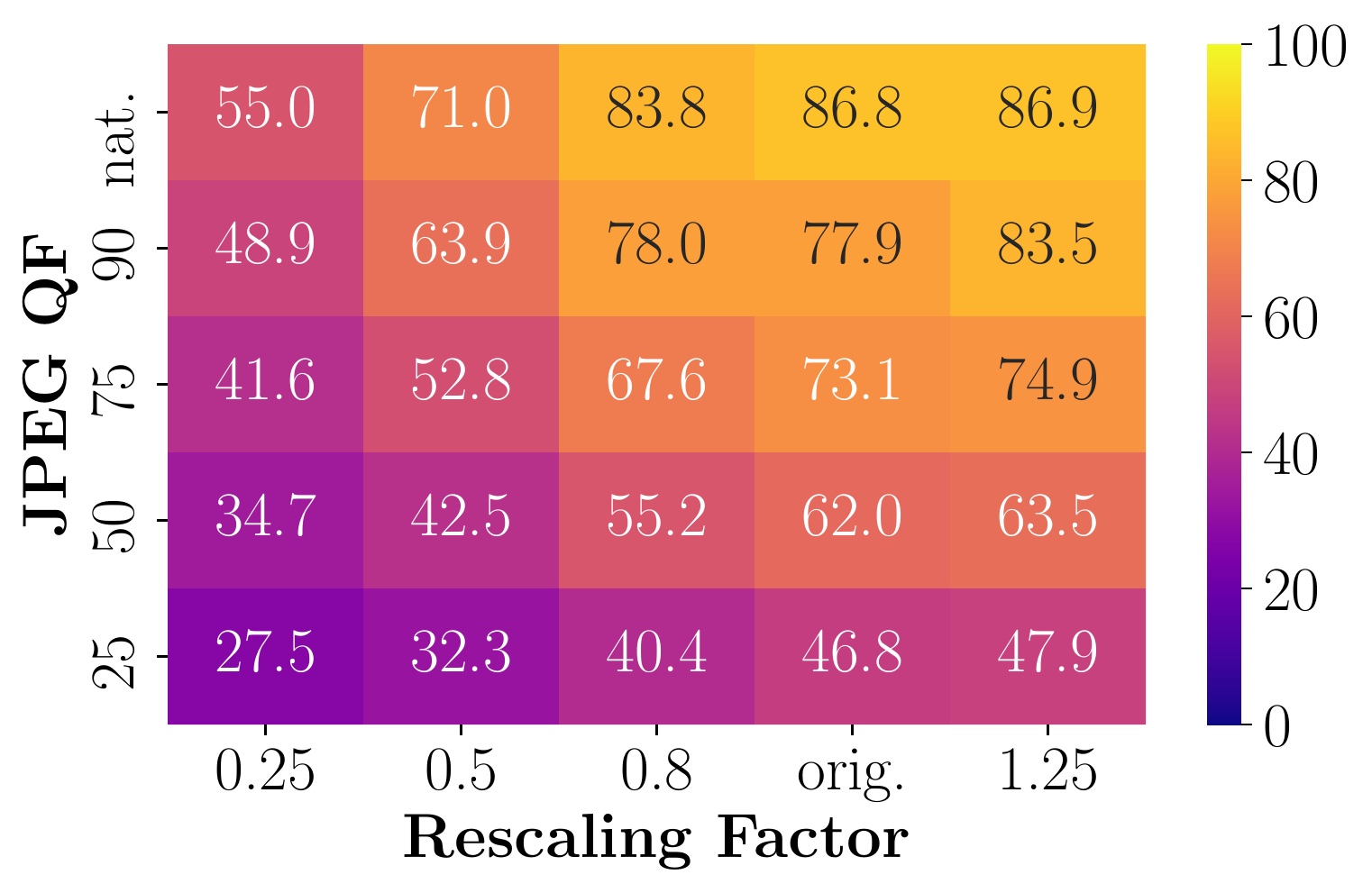}}
	\end{subfloat}
\end{figure*}
\begin{figure*}[!t]
	\vspace{-0.75cm}
	\centering
	\begin{subfloat}{\includegraphics[width=0.245\textwidth]{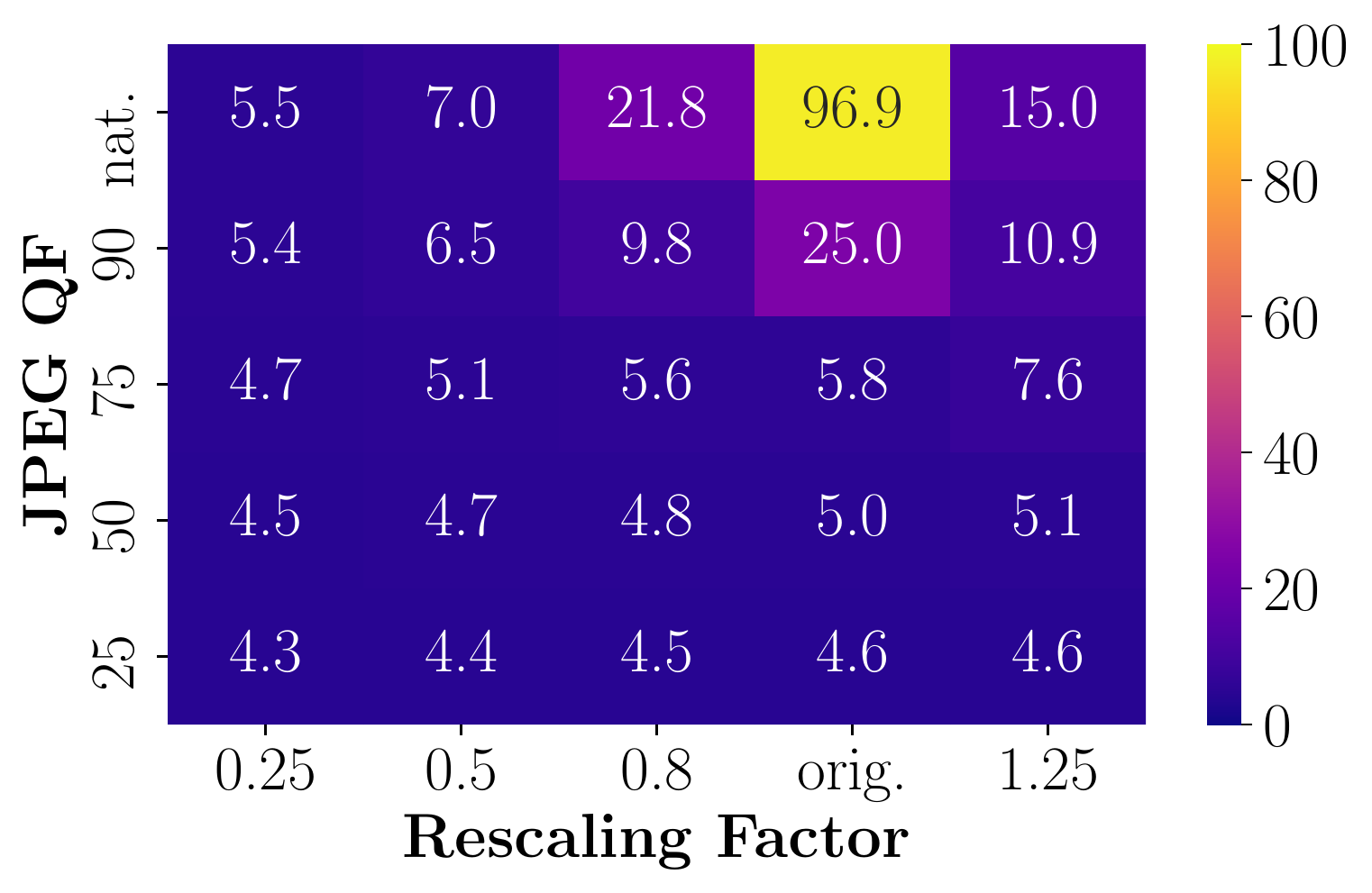}}
	\end{subfloat}
	\begin{subfloat}{\includegraphics[width=0.245\textwidth]{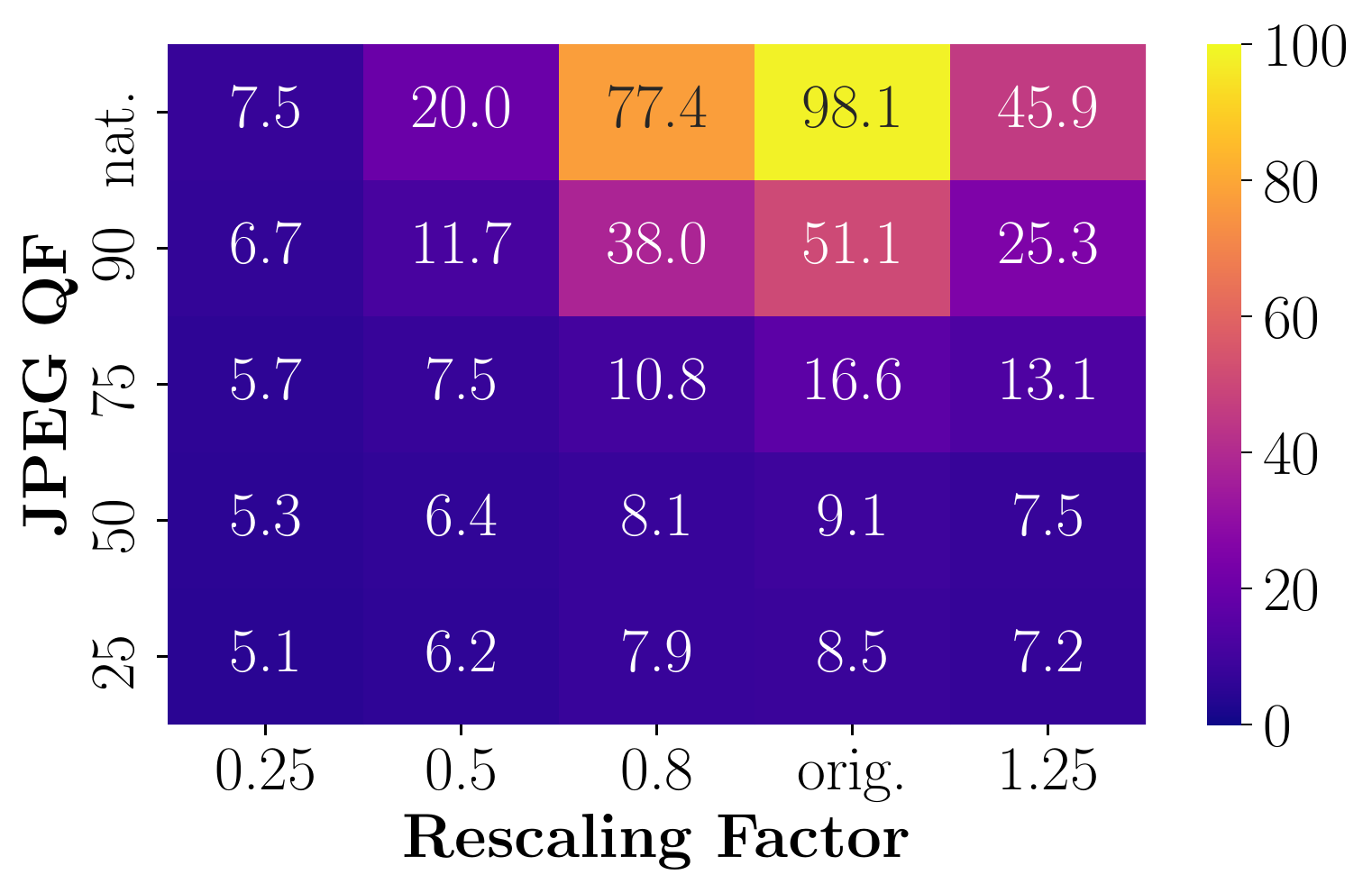}}
	\end{subfloat}
	\begin{subfloat}{\includegraphics[width=0.245\textwidth]{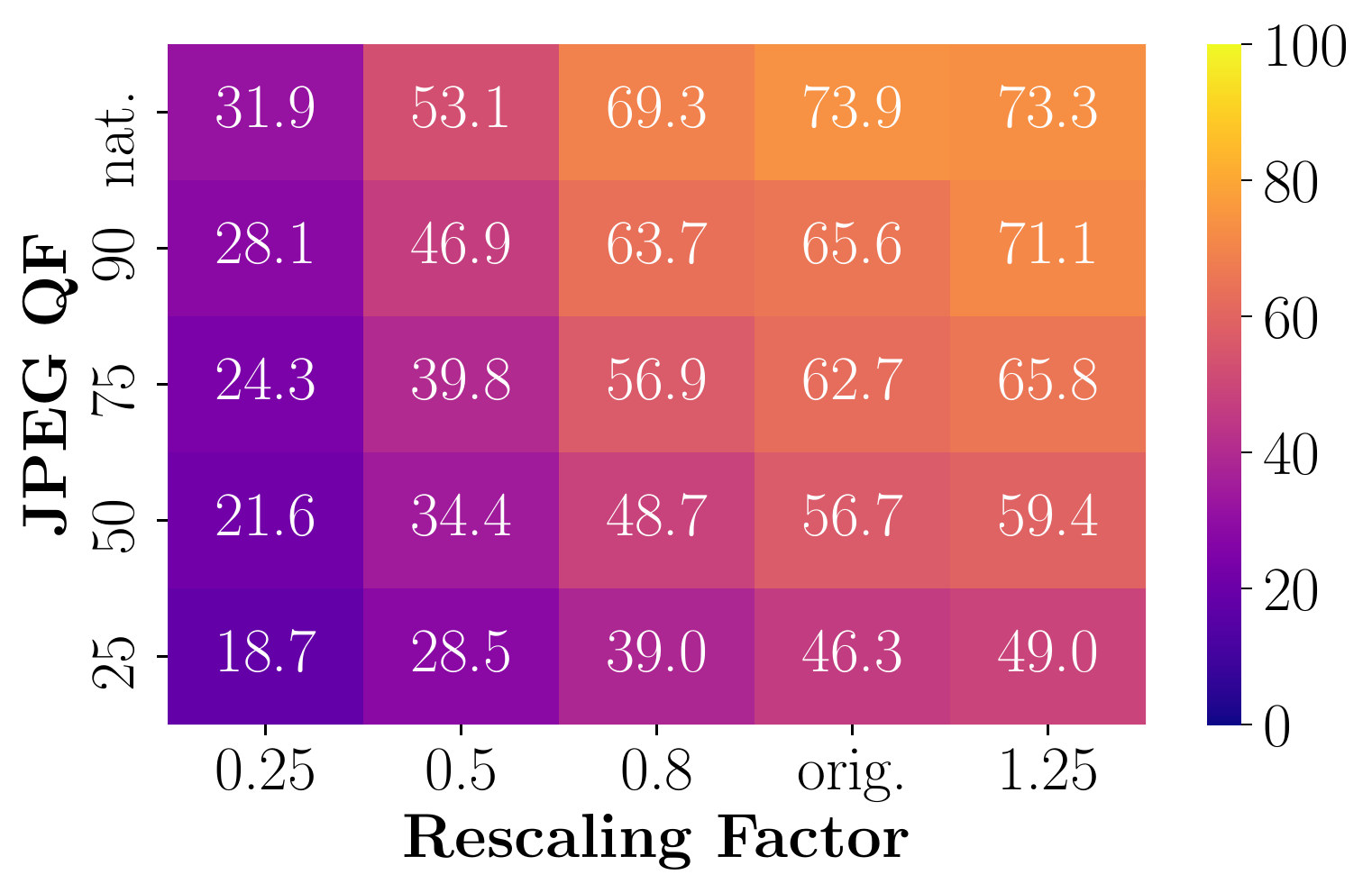}}
	\end{subfloat}
	\begin{subfloat}{\includegraphics[width=0.245\textwidth]{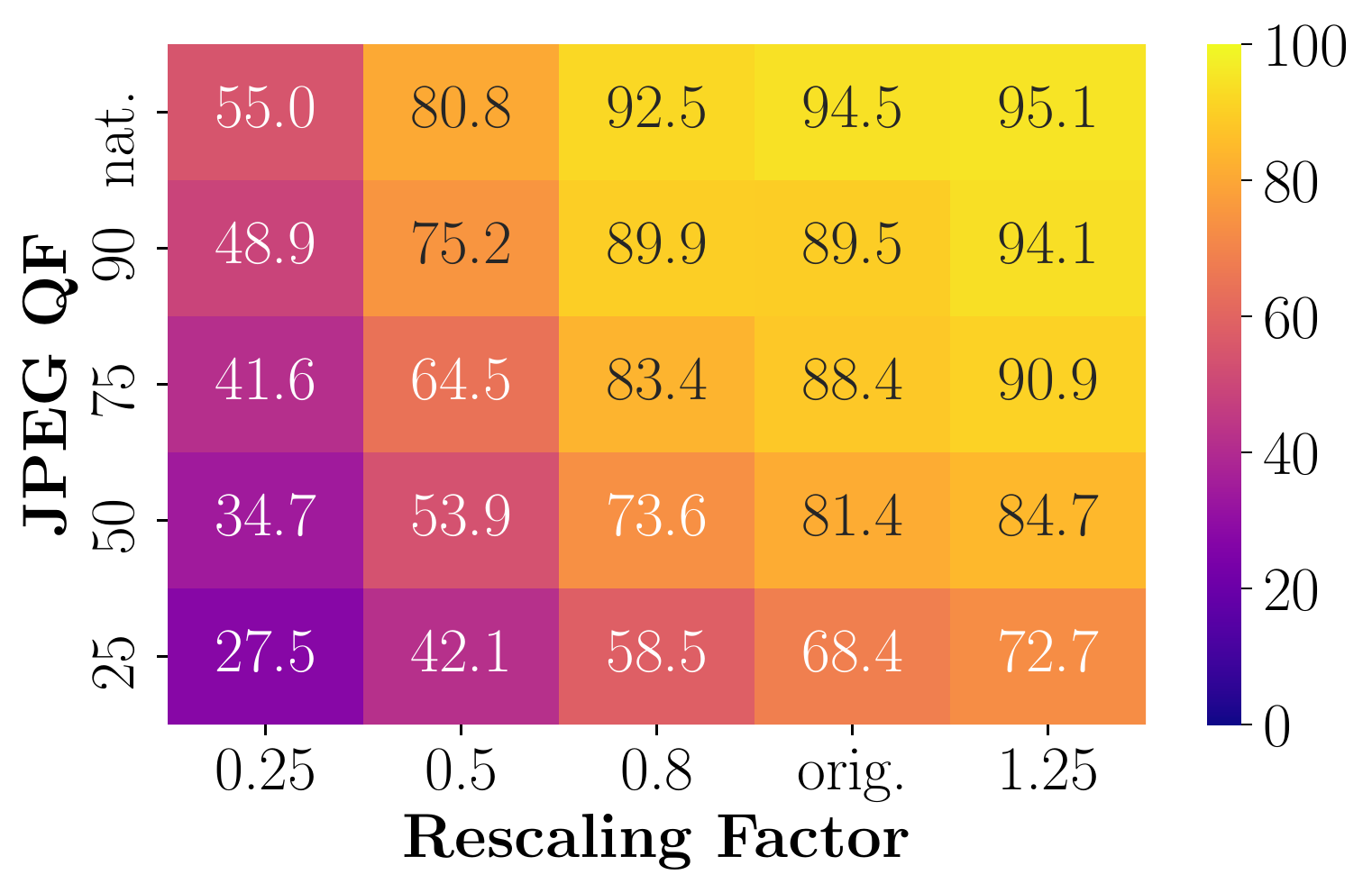}}
	\end{subfloat}
\end{figure*}
\begin{figure*}[!t]
	\vspace{-0.75cm}
	\centering
	\begin{subfloat}[RemNet, no degr.]{\includegraphics[width=0.245\textwidth]{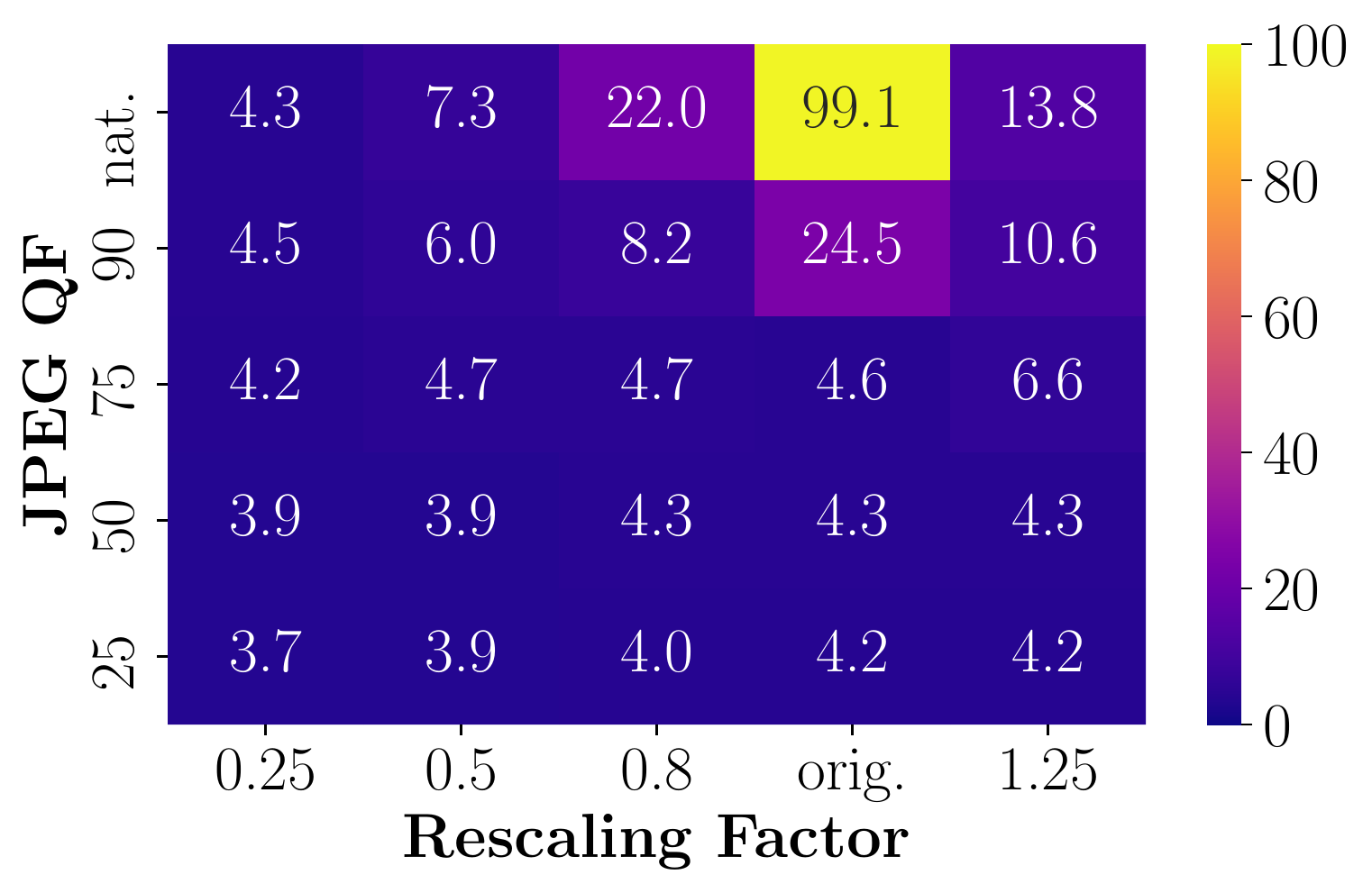}\label{fig:remnet_nodegr}}
	\end{subfloat}
	\begin{subfloat}[EfficientNet-B5, no degr.]{\includegraphics[width=0.245\textwidth]{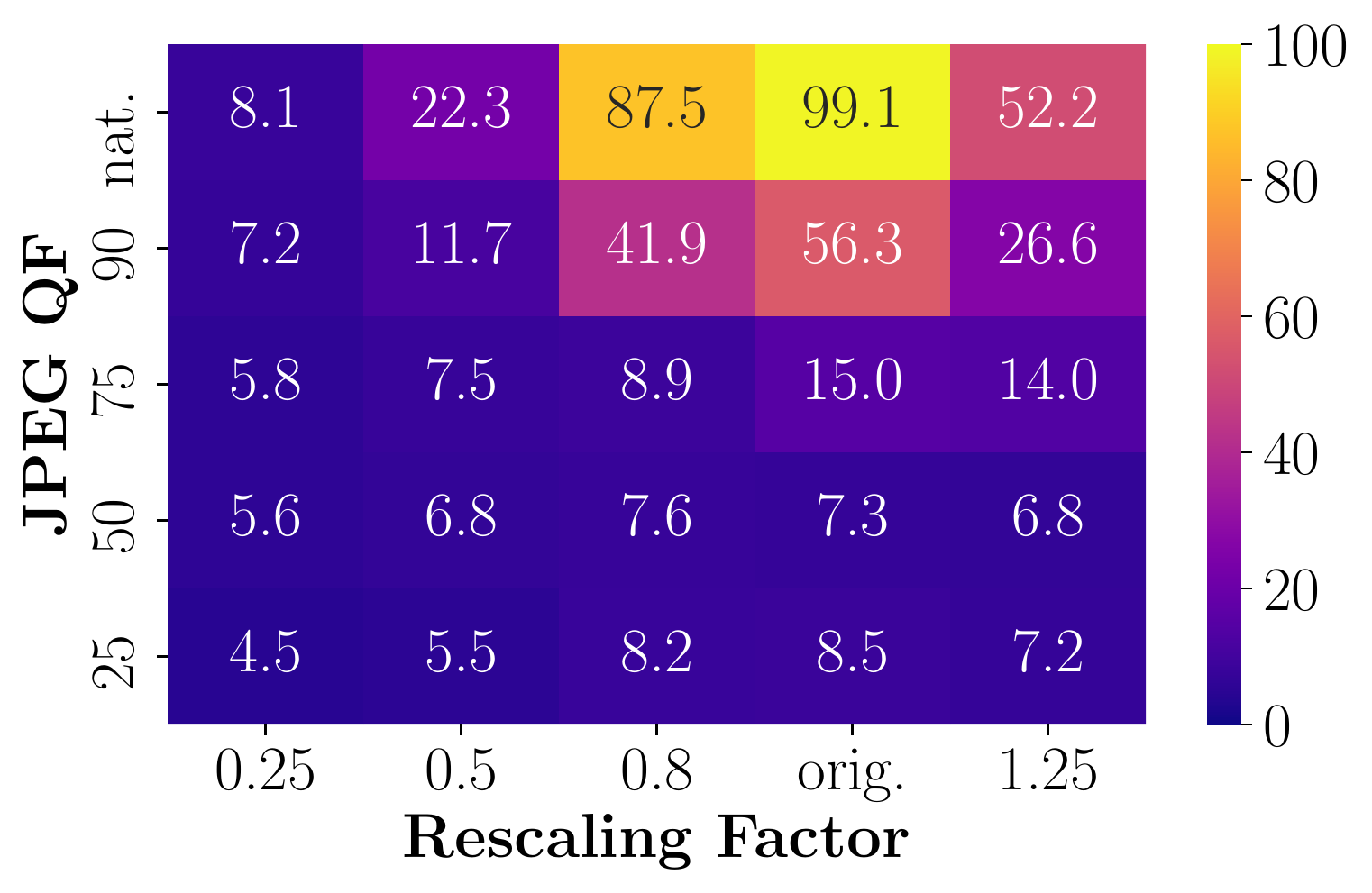}\label{fig:efnb5_nodegr}}
	\end{subfloat}
	\begin{subfloat}[RemNet, degr.]{\includegraphics[width=0.245\textwidth]{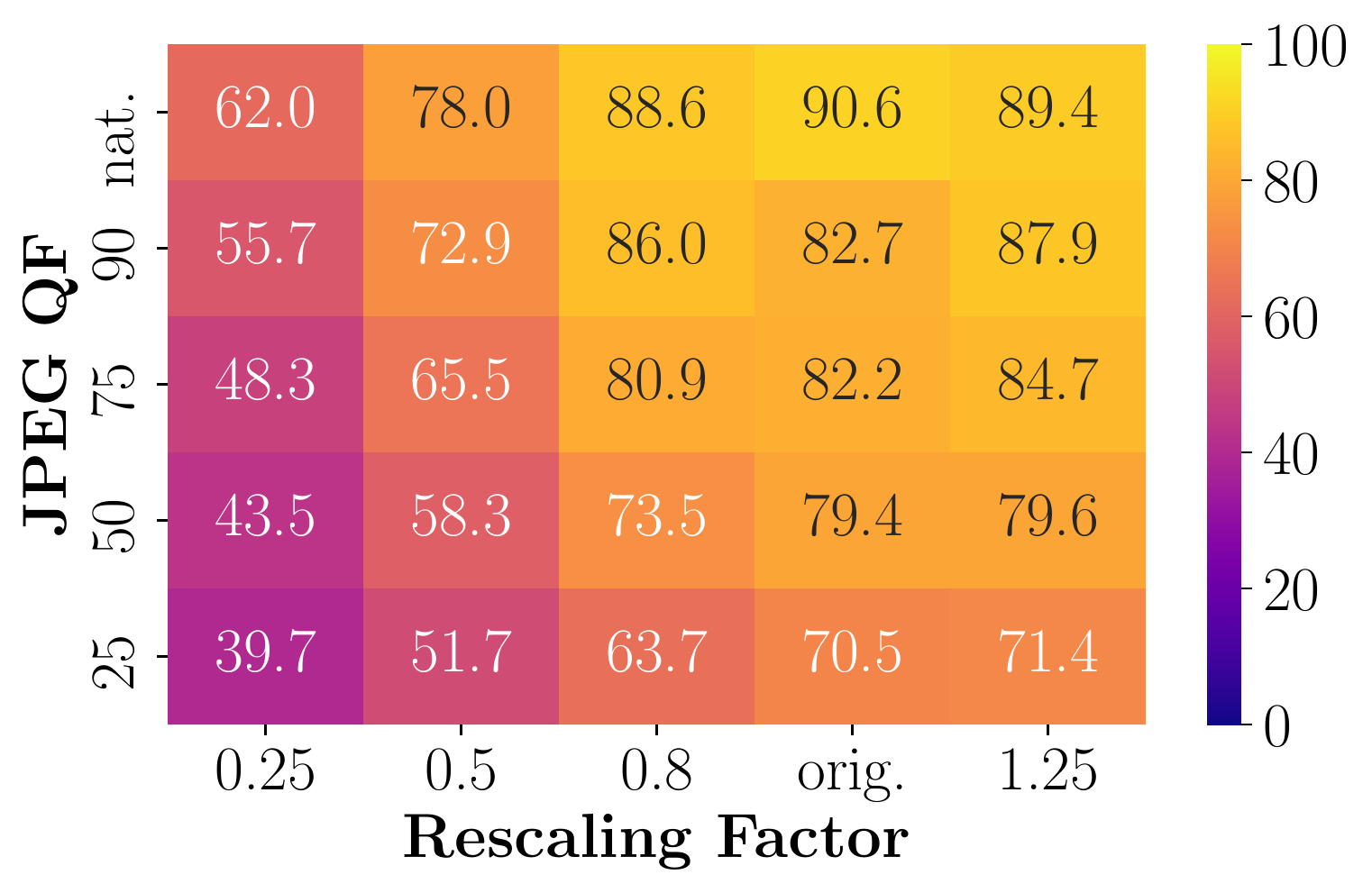}\label{fig:remnet_degr}}
	\end{subfloat}
	\begin{subfloat}[EfficientNet-B5, degr.]{\includegraphics[width=0.245\textwidth]{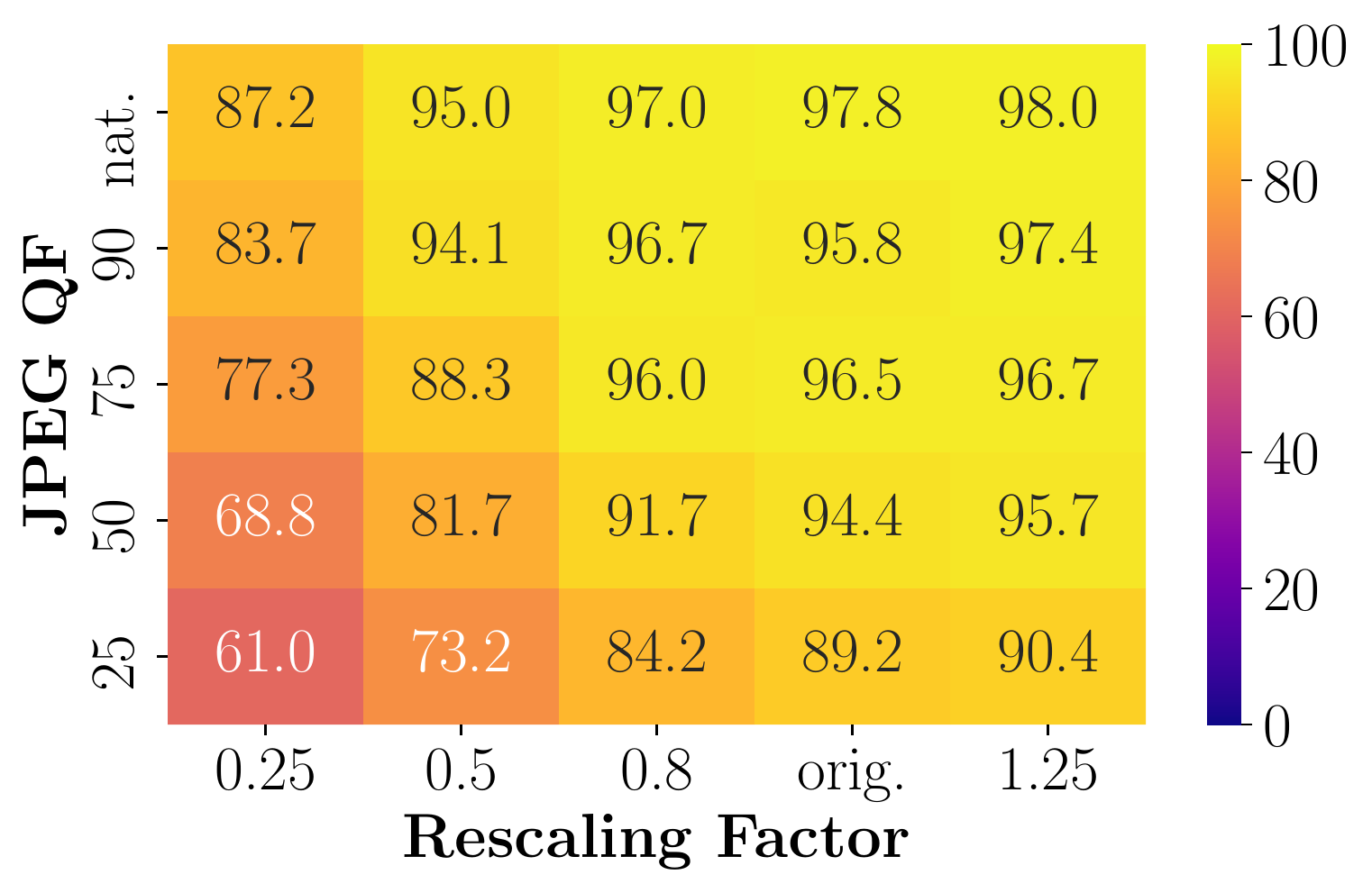}\label{fig:efnb5_degr}}
	\end{subfloat}
	\caption{Robustness against rescaling and JPEG recompression for predictions per patch (first row), cluster (second row) and image (third row).  Columns (a) and (b) show RemNet and EfficientNet-B5 accuracies without augmentation of degradations during training. Columns (c) and (d) show RemNet and EfficientNet-B5 accuracies with augmentation of degradations during training.}
	\label{fig:robustness}
\end{figure*}

\section{Results}
\label{sec:results}

\subsection{Performance Under Ideal Conditions}

In this experiment, we benchmark CNNs for camera identification under ideal
conditions without any post-processing. During training, we only augment with
flipping and rotation, but not with resizing or JPEG recompression.

Table~\ref{tab:fodb_performance_clean} shows the per-patch, per-cluster and
per-image accuracies for the original (camera-native) test images of FODB.
EfficientNet-B5 consistently outperforms the other CNNs for patches and
clusters with accuracies of $96.3\%$ and $98.1\%$, respectively.  For
image-level classification, RemNet and EfficientNet-B5 are on par with an
accuracy of $99.1\%$.  Majority voting improves individual predictions across
all CNNs, which indicates some degree of statistical independence of the
prediction errors.


\begin{table}[t]
	\centering
	\caption{Accuracy (in percent) for closed-set camera identification on camera-native FODB test images.}
	\label{tab:fodb_performance_clean}
	\begin{tabular}{lccc}
		\toprule
		\textbf{CNN} & \textbf{Patch} & \textbf{Cluster} & \textbf{Image}\\\midrule
		BondiNet\cite{bondi2017first} & 71.4 & 84.9 & 93.1\\
		MISLnet~\cite{mayer2019forensic} & - & 93.5 & 96.8\\
		RemNet~\cite{rafi2020remnet} & 93.8 & 96.6 & \textbf{99.1}\\
		EfficientNet-B5~\cite{tan2019efficientnet} & \textbf{96.3} & \textbf{98.1} & \textbf{99.1}\\\bottomrule
	\end{tabular}
\end{table}

\subsection{Robustness Against Known Post-Processing}

In this experiment, we take a closer look at the two best performing CNNs on clean images, RemNet and EfficientNet-B5, and evaluate their robustness against post-processing. For this, we determine the test accuracy on FODB for all combinations of rescaling with factors $f_\mathrm{test} \in\{0.25, 0.5, 0.8, \mathrm{orig.}, 1.25\}$ and JPEG recompression with quality factors $\mathrm{QF}_\mathrm{test} \in \{\mathrm{nat.}, 90, 75, 50, 25\}$.
$\mathrm{orig.}$ (original), resp.\ $\mathrm{nat.}$ (native) indicates no
rescaling and no JPEG recompression. Note that rescaling is applied to patch
clusters prior to patch extraction, which quadratically scales the number of
patches for majority voting on patch clusters and images with
$f_\mathrm{test}$.

Figure~\ref{fig:remnet_nodegr} and Fig.~\ref{fig:efnb5_nodegr} show the accuracies for RemNet and EfficientNet-B5. From top to bottom are accuracies on patch level, cluster level, and image level.
Throughout all qualities, EfficientNet-B5 outperforms RemNet. In most cases,
majority voting again increases the accuracy. While accuracies for both CNNs
are almost perfect for camera-native images ($\mathrm{orig.}$, $\mathrm{nat.}$)
with $99.1\%$, it rapidly decreases on post-processed images.  This is not
surprising, since only high quality images are used for training.  The CNNs
likely rely on fragile high-frequent traces, which are attenuated by
postprocessing~\cite{tsipras2018robustness}.

We retrain both CNNs with artificial degradations as described in Sec.~\ref{sec:augmentation} to
improve the robustness against post-processing. The results for these retrained
CNNs are shown in Fig.~\ref{fig:remnet_degr} and Fig.~\ref{fig:efnb5_degr}.
Already at patch-level, the accuracies of both CNNs are much more stable
compared to training without degradations. For example, at
$f_\mathrm{test}=0.5$, $\mathrm{QF}_\mathrm{test}=75$, the test accuracies at
patch-level amount to $32.4\%$ and $52.8\%$ for both CNNs,
compared to $5.1\%$ and $7.6\%$ without these augmentations.
Moreover, EfficientNet-B5 remarkably outperforms RemNet. For example, the patch-level performance on clean images, is $56.1\%$ for RemNet and $83.8\%$ for EfficientNet-B5. For both CNNs, majority voting further significantly improves the performance.
For image-level decisions and camera-native images, the performance of EfficientNet-B5 trained with degradations ($97.0\%$) is close to testing without degradations ($99.1\%$).
RemNet has difficulties to fully benefit from augmentation with degraded images, with accuracies dropping from $99.1\%$ without degradations to $88.6\%$ with degradations.
We hypothesize that this difference can in part be attributed to the significantly larger capacity of EfficientNet-B5: while both CNNs perform comparably on the easier task of clean images, a CNN with larger capacity might be required for additionally learning the traces of degraded images.
Still, also the superior EfficientNet-B5 shows an accuracy-robustness trade-off, a phenomenon that has been observed for adversarial training before~\cite{tsipras2018robustness, raghunathan2019adversarial}.

\subsection{Robustness Against Unknown Real-World Post-Processing}\label{sec:realworld}

\begin{table*}[!h]
	\centering
	\caption{Test accuracies on FODB for camera-native and post-processed images. Top: boldface shows the best accuracy per column for unknown post-processing. Bottom: blue shows the oracle performance for known post-processing at training.}
	\label{tab:test_accuracies_fodb}
	\begin{tabular}{ccccccccccccccc}
		\toprule
		\multicolumn{3}{c}{}& \multicolumn{12}{c}{\textbf{Test Dataset}}\\\cmidrule{4-15}
		\multicolumn{3}{c}{\textbf{Training Parameters}} &
		\multicolumn{2}{c}{\textbf{orig}} & \multicolumn{2}{c}{\textbf{FB}} & \multicolumn{2}{c}{\textbf{IG}} & \multicolumn{2}{c}{\textbf{TG}} & \multicolumn{2}{c}{\textbf{TW}} & \multicolumn{2}{c}{\textbf{WA}}\\\midrule
		\textbf{CNN} & \textbf{Dataset} & \textbf{Degr.} & \textbf{patch} & \textbf{image} & \textbf{patch} & \textbf{image} & \textbf{patch} & \textbf{image} & \textbf{patch} & \textbf{image} & \textbf{patch} & \textbf{image} & \textbf{patch} & \textbf{image} \\\midrule
		\textbf{RemNet} & orig & no & 93.8 & \textbf{99.1} & 4.0 & 3.6 & 4.2 & 4.2 & 4.5 & 4.3 & 5.5 & 4.7 & 4.2 & 3.9 \\
		\textbf{RemNet} & orig & yes & 59.3 & 90.6 & 18.4 & 36.0 & 22.9 & 48.9 & 26.2 & 52.8 & 37.3 & 74.2 & 24.2 & 50.8 \\
		\textbf{EN-B5} & orig & no & \textbf{96.3} & \textbf{99.1} & 4.9 & 4.6 & 5.7 & 5.6 & 5.7 & 5.3 & 10.8 & 9.8 & 7.0 & 6.8 \\
		\textbf{EN-B5} & orig & yes & 86.5 & 98.0 & \textbf{27.7} & \textbf{51.1} & \textbf{35.4} & \textbf{67.5} & \textbf{42.2} & \textbf{73.1} & \textbf{60.7} & \textbf{93.2} & \textbf{38.5} & \textbf{72.9}\\\midrule
		\textbf{EN-B5} & FB & no & 13.8 & 23.6 & \color{blue}{38.4} & \color{blue}{71.4} & 29.1 & 51.1 & 28.5 & 44.2 & 23.8 & 38.3 & 30.8 & 54.7 \\
		\textbf{EN-B5} & IG & no & 8.1 & 9.4 & 28.4 & 52.1 & \color{blue}{52.1} & \color{blue}{84.0} & 13.5 & 14.0 & 12.1 & 14.0 & 40.4 & 69.1 \\
		\textbf{EN-B5} & TG & no & 16.7 & 23.3 & 21.1 & 32.4 & 25.5 & 37.6 & \color{blue}{57.2} & \color{blue}{86.2} & 35.4 & 55.0 & 32.8 & 51.5  \\   
		\textbf{EN-B5} & TW & no & 36.4 & 57.1 & 14.7 & 21.9 & 25.6 & 41.9 & 28.2 & 41.6 & \color{blue}{76.2} & \color{blue}{97.7} & 33.6 & 54.2 \\ 
		\textbf{EN-B5} & WA & no & 17.3 & 27.3 & 28.8 & 52.4 & 41.5 & 69.2 & 31.0 & 45.3 & 28.4 & 38.3 & \color{blue}{60.0} & \color{blue}{90.4} \\\bottomrule 
	\end{tabular}
\end{table*}

\begin{table*}[!h]
	\centering
	\caption{Test accuracies on VISION for camera-native and post-processed images. Top: boldface shows the best accuracy per column for unknown post-processing. Bottom: blue shows the oracle performance for known post-processing at training. LQ and HQ denote low quality and high quality.}
	\label{tab:test_accuracies_vision}
	\begin{tabular}{ccccccccccc}
		\toprule
		\multicolumn{3}{c}{} & \multicolumn{8}{c}{\textbf{Test Dataset}}\\\cmidrule{4-11}
		\multicolumn{3}{c}{\textbf{Training Parameters}}& \multicolumn{2}{c}{\textbf{orig}} & \multicolumn{2}{c}{\textbf{FB (LQ)}} & \multicolumn{2}{c}{\textbf{FB (HQ)}} & \multicolumn{2}{c}{\textbf{WA}} \\\midrule
		\textbf{Archit.} & \textbf{Dataset} & \textbf{Degr.} & \textbf{patch} & \textbf{image} & \textbf{patch} & \textbf{image} & \textbf{patch} & \textbf{image} & \textbf{patch} & \textbf{image}\\\midrule
		\textbf{RemNet} & orig     & no  & 93.9 & 98.6 & 4.6 & 4.1 & 6.4 & 7.2 & 9.2 & 12.8\\
		\textbf{RemNet} & orig     & yes & 64.7 & 86.7 & 34.8 & 64.0 & 41.5 & 73.9 & 45.7 & 75.9\\
		\textbf{EN-B5}  & orig     & no  & \textbf{95.9} & \textbf{99.2} & 4.9 & 5.9 & 8.3 & 8.7 & 10.2 & 11.8\\
		\textbf{EN-B5}  & orig     & yes & 88.4 & 97.0 & \textbf{46.3} & \textbf{77.0} & \textbf{57.7} & \textbf{88.4} & \textbf{66.5} & \textbf{92.4}\\\midrule
		\textbf{EN-B5}  & FB (LQ) & no  & 8.8 & 14.3 & \color{blue}{64.6} & \color{blue}{88.5} & 22.3 & 33.0 & 28.9 & 40.7\\
		\textbf{EN-B5}  & FB (HQ) & no  & 31.7 & 43.7 & 27.4 & 39.9 & \color{blue}{72.8} & \color{blue}{95.4} & 25.4 & 36.8\\
		\textbf{EN-B5}  & WA      & no  & 21.6 & 32.9 & 30.5 & 47.5 & 18.3 & 27.8 & \color{blue}{77.3} & \color{blue}{96.3}  \\\bottomrule
	\end{tabular}
\end{table*}

\begin{table*}[!h]
	\centering
	\caption{Test accuracies for splits on sorted images of VISION for camera-native and post-processed images. Top: boldface indicates the best accuracy per column for unknown post-processing. Bottom: blue indicates the oracle performance for known post-processing. LQ and HQ denote low quality and high quality.}
	\label{tab:test_accuracies_vision_timesorted}
	\begin{tabular}{ccccccccccc}
		\toprule
		\multicolumn{3}{c}{} & \multicolumn{8}{c}{\textbf{Test Dataset}} \\\cmidrule{4-11}
		\multicolumn{3}{c}{\textbf{Training Parameters}} & \multicolumn{2}{c}{\textbf{orig}} & \multicolumn{2}{c}{\textbf{FB (LQ)}} & \multicolumn{2}{c}{\textbf{FB (HQ)}} & \multicolumn{2}{c}{\textbf{WA}} \\\midrule
		\textbf{Archit.} & \textbf{Dataset} & \textbf{Degr.} & \textbf{patch} & \textbf{image} & \textbf{patch} & \textbf{image} & \textbf{patch} & \textbf{image} & \textbf{patch} & \textbf{image}\\\midrule
		
		\textbf{RemNet} & orig   & no  &  87.5 & 93.2 & 3.7 & 4.8 & 4.7 & 5.5 & 5.5 & 7.4\\
		
		\textbf{RemNet} & orig   & yes &  44.2 & 67.9 & 20.1 & 39.5 & 25.8 & 52.4 & 27.9 & 51.7 \\
		
		\textbf{EN-B5} & orig    & no   & \textbf{87.8} & \textbf{93.9} & 3.6 & 4.5 & 7.0 & 7.4 & 6.9 & 8.6\\
		
		\textbf{EN-B5} & orig    & yes  & 76.8 & 88.5 & \textbf{28.2} & \textbf{54.1} & \textbf{40.9} & \textbf{70.5} & \textbf{44.7} & \textbf{72.4} \\\midrule
		
		\textbf{EN-B5} & FB (LQ) & no   & 7.3 & 10.8 & \color{blue}{42.3} & \color{blue}{67.6} & 16.8 & 26.3 & 20.7 & 29.1\\  
		
		\textbf{EN-B5} & FB (HQ) & no   & 26.3 & 39.0 & 18.0 & 26.0 & \color{blue}{55.2} & \color{blue}{83.7} & 16.1 & 21.6\\		 
		
		\textbf{EN-B5} & WA      & no   & 18.7 & 30.7 & 19.4 & 30.8 & 14.4 & 21.8 & \color{blue}{56.1} & \color{blue}{82.2}\\\bottomrule
	\end{tabular}
\end{table*}


In this section, we evaluate the robustness of RemNet and Efficient\-Net-B5 against
real-world post-processing by unknown algorithms and parameters, as it occurs
during social network sharing.  We again train both CNNs once without and once
with degradations. The networks do not obtain examples of social media images
for training.

We evaluate the selected models on original images and all five post-processed
versions of the test images (Facebook: FB, Instagram: IG, Telegram: TG,
Twitter: TW, Whatsapp: WA). The resulting accuracies are listed in
Tab.~\ref{tab:test_accuracies_fodb}.
When training without degradations, the networks can only excell on original
images, analogously to the previous experiments.
Pretraining on DIDB slightly improves the performance of EfficientNet-B5 on
clean images.
Augmentation with artificial degradations significantly improves the
performance of both CNNs on all social network data, eventhough social media
data itself was not part of the training.
Again, EfficientNet-B5 largely outperforms RemNet in all experiments.

We perform an additional experiment as a reference for the impact of prior
knowledge on the data: we pretrained EfficientNet-B5 on DIDB with
degradations. Additionally, we feed the social network images from the \emph{training} set
to EfficientNet-B5 as an oracle for the \emph{test} set degradations.
Table~\ref{tab:test_accuracies_fodb} (bottom) shows that such strong prior
knowledge yields at image level accuracy gains from  $4.5~\mathrm{p.p.}$ for Twitter
(with baseline already $93.2\%$) up to $20.3~\mathrm{p.p.}$ for Facebook.

\subsection{Impact of Scene Splitting}\label{sec:splitting}
\label{sec:scene_splitting}

In this experiment, we analyze the influence of scene splitting on CNN-based
camera identification on the VISION dataset. The scene content is not
constrained in several datasets including VISION which prevents splitting by scenes.
Some per-device image sets in VISION are highly correlated, such that
randomized splitting makes it likely that training and test sets contain images
of identical cameras with similar content. We conjecture that scene content may
open a side-channel that CNNs are prone to exploit, which may lead to an
overestimation of the CNN generalization.
We show empirical evidence for this conjecture in two experiments.

First, we randomly split the VISION images in training, validation and test
sets. We use the evaluation protocol by Marra~\textit{et
al.}~\cite{marra2018vulnerability} and use the $29$ unique devices with random
guessing accuracy of $29^{-1} = 3.45\%$.

Second, we make an attempt to improve the splitting strategy, and to further
separate image content between training and test set. 
To this end, we sort the images of each device by their acquisition time using \texttt{DateTimeOriginal} from the EXIF file header, and split the dataset  along the timeline of the images.
In this way, similar images recorded within a short period of time are likely
to be either in the training or test set, but not in both. This significantly
reduces overlap in image content between training and test set.
Except of the splitting policy, all settings are identical between both
experiments.

Results for the first and second experiment are shown in Tab.~\ref{tab:test_accuracies_vision} and Tab.~\ref{tab:test_accuracies_vision_timesorted}.
Performances drop significantly when moving from completely random splits (Tab.~\ref{tab:test_accuracies_vision}) to splits by timestamp (Tab.~\ref{tab:test_accuracies_vision_timesorted}). For example, on clean images the accuracy of EfficientNet-B5 without degradation drops from $99.2\%$ to $93.9\%$. The performance of EfficientNet-B5 with degradation for Whatsapp-compressed test images drops even by $20~\mathrm{p.p.}$, from $92.4\%$ to $72.4\%$.
This discrepancy suggests that scene content contributes to the results in Tab.~\ref{tab:test_accuracies_vision}.
Moreover, such a side-channel may prevent the CNN from learning more relevant traces.
We hence believe that the results in Tab.~\ref{tab:test_accuracies_vision_timesorted} are closer to the performance that can be expected in practice.
These observations emphasize the importance of a rigorous scene splitting as supported by FODB.

\section{Conclusion}
\label{sec:conclusion}

This work proposes the Forchheim Image Database (FODB) as a new benchmark for image forensics algorithms under real-world post-processing.
Our database consists of more than 23,000 images of 143 scenes by 27 smartphone
devices of 25 models and 9 brands. FODB combines clean
training/validation/test data splits by scene with a wide range of modern
smartphone devices shared through a total of five social network sites, which
allows rigorous evaluations of forensic algorithms on real-world image
distortions.  We demonstrate FODB's usefulness in an evaluation on the task of
camera identification. Our results provide three insights. First, the
general-purpose network EfficientNet-B5 largely outperforms three specialized
CNNs. Second, EfficientNet-B5's large capacity also fully benefits from
training data augmentation to generalize to unseen degradations. Third, 
clean data splits by scenes can help to better predict generalization
performance.







\bibliographystyle{IEEEtran}
%
\bibliography{references.bib}

\end{document}